# A Global Multi-Unit Calibration as a Method for Large Scale IoT Particulate Matter Monitoring Systems Deployments

Saverio De Vito, Gerardo D'Elia, Sergio Ferlito, Girolamo Di Francia
Miloš Davidović, Duška Kleut, Danka Stojanović, Milena Jovašević-Stojanović

*Abstract—* **Scalable and effective calibration is a fundamental requirement for Low Cost Air Quality Monitoring Systems and will enable accurate and pervasive monitoring in cities. Suffering from environmental interferences and fabrication variance, these devices need to encompass sensors specific and complex calibration processes for reaching a sufficient accuracy to be deployed as indicative measurement devices in Air Quality (AQ) monitoring networks. Concept and sensor drift often force calibration process to be frequently repeated. These issues lead to unbearable calibration costs which denies their massive deployment when accuracy is a concern. In this work, We propose a zero transfer samples, global calibration methodology as a technological enabler for IoT AQ multisensory devices which relies on low cost Particulate Matter (PM) sensors. This methodology is based on field recorded responses from a limited number of IoT AQ multisensors units and machine learning concepts and can be universally applied to all units of the same type. A multi season test campaign shown that, when applied to different sensors, this methodology performances match those of state of the art methodology which requires to derive different calibration parameters for each different unit. If confirmed, these results show that, when properly derived, a global calibration law can be exploited for a large number of networked devices with dramatic cost reduction eventually allowing massive deployment of accurate IoT AQ monitoring devices. Furthermore, this calibration model could be easily embedded on board of the device or implemented on the edge allowing immediate access to accurate readings for personal exposure monitor applications as well as reducing long range data transfer needs.**

*Index Terms—* **IoT Air Quality Monitoring Devices, IoT Air Quality Sensors Accuracy, Multi-Unit Scalable Calibration, Field Calibration, Embedded Machine Learning.**

This work was partially funded by EU through UIA Air-Heritage and H2020 VIDIS Project under grant agreement No 952433. We are also grateful to Ministry of Science, Technological Development and Innovations of Serbia under themes 1002201 and 0402312 which are realized in Vinca Institute of Nuclear Sciences (contract no. 451-03-47/2023-01/ 200017).

(Corresponding author: S. De Vito – saverio.devito@enea.it).

Saverio De Vito, Gerardo D'Elia, Sergio Ferlito and Girolamo Di Francia are with ENEA - Agenzia per le Nuove Tecnologie, l' Energia e lo Sviluppo Economico Sostenibile, Photovoltaics and Sensors Applications and Systems Lab., Smart Devices Division, Energy Technology Dept., C.R. Portici, P.le E. Fermi, 1 – 80055 Portici (Naples) – Italy (e-mail: name.surname@enea.it).

M. Davidovic, D. Kleut, D. Stojanovic and M. Jovasevic-Stojanovic are with Vinča Institute of Nuclear Sciences, National Institute of the Republic of Serbia, University of Belgrade, Mike Petrovića Alasa 12-14, 11351 Vinca, Belgrade, Serbia.

## I. INTRODUCTION

Particulate matter concentrations are considered among the primary drivers of air quality related concerns worldwide. Epidemiological studies have demonstrated their fundamental role for human health which include increasing the incidence of cardiovascular diseases and several types of cancer [1] [2]. Achieving a clear, and objective assessment of pollutant levels is hence necessary for informing citizens about the safety of the urban environment the possible outcomes fo their present and future health status. *A fortiori*, monitoring this phenomena with accurate, reliable, methods is paramount to continuously assess AQ levels, and has deep legal implications for protecting the citizens and validating the impact of remediation actions, eventually generating political accountability ground. Low cost air quality monitor systems (LCAQMS), based on solid state gas, particle sensors and IoT paradigm are increasingly penetrating the AQ monitoring market due to their capability to increase the spatial and temporal density of AQ information in most countries [32]. When integrated within ad-hoc IoT device management and GIS software they can contribute high valued hyper-local information about air quality [33]. In facts, the gains in spatial resolution, achievable by the deployment of a network of these devices, is considerable [3] when compared with the very accurate but costly, bulky and, ultimately, sparse network of regulatory grade analyzers. They ultimately allow for designing and implementing pinpoint remediation policies as well as discover and highlight environmental inequalities in urban centers, even in low GDP countries. An example of LCAQMS is the MONICA™ device, a microcontroller based, smartphone interfacing, battery operated architecture developed by ENEA, featuring gas and particulate mobile monitoring capabilities being based on low cost electrochemical gas sensors and optical particle counter sensor. Unfortunately, LCAQMS accuracy is hampered by several issues including environmental conditions interference. This is limiting their deployment whenever accuracy is of concern as when IoT devices are to be integrated in hierarchical, regulatory monitoring networks [20]. Recently, Maidan et al. [34] discussed the requirements for their effective deployment putting sensors or so called "virtual sensors" calibration as a basic enabler at the core of their proposed architecture. They recognize that pursuing an increase in data quality and data availability for LCAQMS is paramount to push the boundaries of administration uptake and citizenship acceptability in different applications. This will allow to turn the gathered data into actionable information enabling the full exploitation of their potential for public health and safety. Several technologies for ameliorating data availability have been implemented relying on statistical data imputation techniques. More recently, the inherently spatial structure of the air pollution process has been exploited to improve the accuracy of data imputation based on graph signal processing paradigm [4]. Given the significant number of sensors usually included in these devices for a comprehensive assessment and, consequently, the amount of raw data generated, the capability of locally obtaining calibrated concentrations assessment, whether on board or on the edge, is indeed very relevant for relaxing data throughput requirements [5]. The latter will further allow a reduction of power needs, extending battery operation time, and allowing for immediate feedback on pollutant exposure thus enabling their use as wearable/mobile personal exposure monitoring devices. Bridging IoT and Artificial Intelligence, field calibration derivation is actually a fundamental step for implementing the LCAQMS data processing pipeline enabling to turn raw data into concentration values for the monitored pollutants [34]. This data fusion technique exploits target pollutant raw sensors data and data related to observable interferents which typically include other pollutants and environmental conditions for interference correction [5]. A widely studied class of low cost sensors are PM sensors for which targets are size-partitioned particle concentration assessment. Commonly based on optical particle counter principles, many of them exploit Mie scattering law to express estimations of the count of particles in a certain size range (i.e. <1um, <2.5um, < 5 um, <10um). They have been used in several embodiments for air quality monitoring purposes being rather successful [33]. Their impressive low cost (around 20 US $ in 2023 for a single PM sensor device, about 300$ for a complete IoT device and Data management service solution) comes with relevant limitation that must be taken into account when accuracy is a concern. High humidity levels negatively affect the size to mass conversion accuracy by causing thin layers of water to wrap around the particle ultimately determining a physical swelling of particles dimensions and changing its overall density. Particles chemical composition also affect density and may change in different emissions scenarios. Furthermore, while fundamentally based on the same principles as lab grade instruments, LCAQMS lack some of the advanced features of their higher priced counterparts, such as precision controlled air pumps (using turbulent fans instead), or protective clean sheath air enveloping ambient air sample thus being more susceptible to aging and harsh environmental conditions. Vendor based calibration relying on fixed size to mass conversion hypothesis may easily become inapplicable in these conditions, determining unacceptable loss of accuracy [6] [7]. As such, advanced post-selling calibration law derivation usually take into account particles count along with humidity assessments as a source of relevant information [8]. Field calibration is the state of the art technique for delivering accurate concentrations estimations for LCAQMS [9]. Differently from lab based calibration in which pollutant mix concentration and environmental conditions are carefully controlled, in this methodology open air colocation with reference analyzers allows to build a dataset joining raw sensors data with 'true' concentrations in purely uncontrolled conditions. This dataset can be exploited to tune parameters of physically rooted/black box sensors models deriving the calibration law enabling to convert raw data into concentration estimations. Fabrication variance generating slightly different sensors properties in the different units, severely harms calibration accuracy when using a single calibration model, sic et simpliciter, for all the sold sensors units. In most of the cases fabrication variance requires the derivation of ad-hoc sensor specific calibrations. Several researchers have further clarified how field calibration derived in reasonably long colocation periods (i.e. several weeks) lacks generalization properties [10] [11]. Actually, when operating conditions change from those encountered during training set recording, the derived calibration law suffers from lack of accuracy. This condition is known as concept drift [12]. It frequently emerges, for example, in long term multi-seasonal deployments or due to relocation of the LCAQMS after calibration [13]. Hence, still far from being scalable, plain Field Calibration would require huge logistics efforts for extending colocation periods for months [14], repeatedly re-calibrate or combining both these approaches [15], for each of the sensors [16]. This easily become unfeasible for both modern commercial IoT devices market and research projects, now needing the deployment of hundreds of accurate sensors for achieving their intended goals [17]. Recently, calibration transfer has been proposed as a way to reduce the data gathering efforts by reducing the number of samples needed to obtain a viable calibration. This methodology relies on the calibration of a single master instrument with lab or field based data and to reduce fabrication variance by processing the raw data response of different "slave" devices. The difference in the response due to different sensors characteristics (sensitivity, response to zero) are in fact strongly reduced, ideally eliminated, by discovering and applying linear transformations in the multivariate response data domain mapping the slave device responses on the master ones. The transformations are found by exploiting the difference observed in the response to predetermined conditions (i.e. at predetermined response drivers values). Transfer learning, based on deep neural networks paradigms is a competing approach that may have a profound impact in the near future [18]. Though reducing the number of single LCAQMS related samples, this methodological framework still needs each sensor colocation with reference instruments in predetermined conditions for a sufficient amount of time.

In search of a scalable calibration methodology for low cost connected devices, Mailings and colleagues [19], have shown the possibility of applying multi-unit additive calibration to several different electrochemical gas sensors in air quality scenario. The calibration is obtained through using datasets considering the median response of several units to identical field recorded conditions. This median calibration showed valuable for calibration of this class of gas sensors by reducing the impacts of fabrication variance obtaining a significant increase in the scalability of the proposed methodology.

Remote calibration [20], exploiting nearby Ozone monitoring reference stations data for updating calibration parameters, is a promising methodology to deal both with fabrication variance as well as with sensors and concept drifts at a limited cost.

Each single device can be repeatedly recalibrated relying on the ecosystem networking infrastructure, often exploiting LoRaWAN, and remote computing infrastructures, by fusing verbose raw data from the device and nearest reference station data stream. However, if reference station deployment conditions (e.g. the distance from emissions sites like main roads, etc.) radically differs from the ones of the node under calibration, plain remote data may provide very limited information. Furthermore, this model implementation increases data transfer needs under limited bandwidth availability conditions. The limited spatial variability of Ozone has been also exploited by Barcelo-Ordinas et al. [21], who have used an interesting Kriging spatial interpolation to impute statistical moments at calibrated LCAQMS position further correcting their developing bias, thus proposing an interesting crosslink between geostatistics and IoT based pervasive measuring approaches. In this work, **the main goal is to propose a simple and scalable universal calibration procedure for LCAQMS and compare it with state-of-the-art ad-hoc field calibration methodology**. Specifically designed for optical particle sizers/counters and nephelometers it relies on field recorded data from a limited number of units. We exploit their expected limited fabrication variance to obtain a universal calibration law which may applied to all units operating in similar conditions. We show how, for low cost particle sensors, this calibration method is sufficient to **obtain similar performance** with respect to ad-hoc sensors field calibration procedures at a **tiny fraction of its cost.**

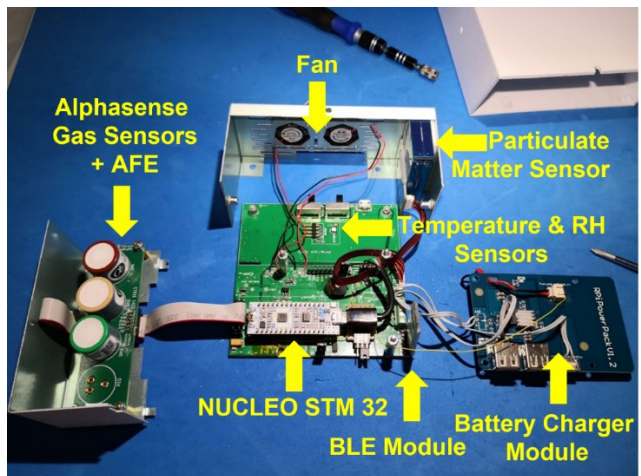

*Figure 1: MONICA(TM) device open box picture.*

## II. RELATED WORKS

As stated above, field calibration is a compulsory step to gain satisfactory data quality level in low-cost sensors technology over medium term [22]. Its general applicability, can exceed targeted gas sensor calibration enabling so called Virtual Sensors capable to obtain high resolution information even when low cost sensors for that specific pollutant are not available [34]. Nevertheless, for large deployments, especially in future smart city applications, field calibration represents a non-negligible cost. One of the emerging approaches to achieve economic viability is the assessment and evaluation of generalized models. In this section we report some of the most promising works in this area. A generalized model can be assimilated to the model of a virtual sensor that can be applied effectively to all copies of the same sensor. However, since two exactly "identical" items do not exist in the reality of a sensor manufacturing process, achieving such a generalized model is not straightforward. The approach of Malings et al [19] attempts to reduce the effects of sensor-to-sensor variability by exploiting the overall information gathering process during a large sensors unit field co-location. They build a training set employing the median between the raw response values obtained by several electrochemical gas sensors simultaneously collocated with a reference station. This training set provided better generalization properties reducing the inherent variability between the instruments. Another interesting property of their median generalized model is the enhanced robustness against the relocation to other sites and ageing. These promising results are obtained with electrochemical gas sensors while in this work we will explore whether similar characteristics are also extensible to PM sensors [22]. Smith and colleagues, used a similar approach to calibrate the median signal of an IoT AQ multisensory based on an array of electrochemical sensors replicas to estimate pollutant concentrations [23]. The node have shown a reduction of uncertainty when estimating linear regression parameter when increasing the number of the sensors participating in the median computation. Furthermore the authors reported a significant improvement in terms of accuracy with respect to individual factory based calibration. However, featuring 6 sensors for each of the predefined targets, their device has an increased cost as well as power consumption though the latter increase was found to be less significant. Ultimately, this approach develop a calibration of a single specific device based on several sensors replica, our goal is instead to obtain a **universal calibration** to be applied to different devices with a single sensor node for targeting cost effectiveness of the overall procedure. Among regression techniques, the Partial Least Squares (PLS) is very successful for dealing with high dimensionality data. Orthogonal-PLS [24] highlights hidden data by finding a relationship between the so-called latent variables which maximize the relationship between them and the target variable. This suggests that there could be robust features which could be insensitive to the fabrication variance. In facts, a different approach based on calibration transfer, was proposed by Solórzano et al. which tried to exploit the common hidden structure between the responses of multiple units of the same sensor model and the gas concentrations [25]. Finding and exploiting this common structure means to discover a model that is suitable for all sensors. While Solorzano et al. apply the procedure to classification problems, Miquel-Ibarz et al. extended its usage to a regression task which is relevant for the scope of this work [26].

## III. THE IoT MONICA DEVICE AND ECOSYSTEM

The data used for the present work have been recorded using a set of 30 different ENEA MONICA devices (see Fig. 1) [9]. MONICA™ is an Italian acronym for Cooperative Air Quality Monitoring since the device was originally devised for citizen science operations as well for mobile use on drones and cars/buses. The device is based on an array of 3 electrochemical sensors (Alphasense A4-NO2, A4-CO, A4-O3 sensors respectively targeted to NO2, CO, O3) and one Plantower 7003 Optical particle counter targeted to PM1, PM2.5 and PM10 concentration estimation, on which this work is focused. The featured gas sensors raw output consist in two readings representing the electrical potential at working and auxiliary electrodes reflecting respectively the concentration of target gas and due temperature correction factor. In facts, these sensors are prone to both temperature and non target gas interference [19]. Details on vendor recommended operating procedures can be found at [35].

The addition of a PM sensor represent an update with respect to the work presented in [9] significantly improving the versatility of the tool. Raw sensors data including size partitioned particle counts and PM concentration estimation along with environmental conditions variables (T,RH) are sampled each 6 seconds. A single PM sensor device generates 6 dimensionally fractionated particulate number concentrations readings (0.3,0.5,1,2.5,5,10 [ug/m^3]) as raw data along with 2 different (named *standard particles* and *atmospheric environment*, respectively) vendor based calibrated values for each of the 3 relevant dimensional fractions (1,2.5,10 [ug/m^3]) for a total of 12 double sized floating point numbers. Lithium Polymer (Li-Po) battery (3000mAh) and Bluetooth™ Low Energy (BLE, *Serial Port Profile*) board allow the system to operate in mobile and autonomous deployment scenarios. BLE interface allows for easy short range connectivity with all recent smart phones platforms while achieving remarkably low energy demand (<15mA peak demand) and sufficient OTA throughput (in our case 500Kbit/sec). Li-Po batteries capacity practically allow for 8-hours of continuous monitoring schedule so to level the need for recharging to the one of the companion smartphone.

Conversely, during colocation experiments with reference analyzers, we used a SBC (Single Board Computer) driven data sink (Raspberry Pi 4, Model B 4 GB - Arm Cortex-A72, 1,50 GHz, 4 GB DDR4, WLAN-AC, Bluetooth 5.0 with RaspbianOS) to receive units data on its BLE interface and forward the extracted JSON encoded data to remote IoT cloud backend systems. The resulting data stream include reference station true data for calibration purposes. It is operated by ARPAC[1] and relies on regulatory grade analyzers for hourly PM concentration estimations. Cloud-side, an NGINX™ server wraps a Node.js engine running the server side routines implementing the logic of RESTful APIs.

The whole IoT backend architecture is actually made up of:

- NGINX™ component, acting as a reverse proxy server (with load balancing features), it is used to publicly expose the backend services (REST API) in a secure and efficient way.

- NodeJS component, the Javascript runtime environment which is used to build the REST API service (i.e. the backend services, coded using ExpressJS as Web Server) that allows to perform CRUD (Create, Read, Update & Delete) operation on Database and data analytics functionalities.

- MongoDB™ [27] which is the selected NoSQL database used to permanently store data coming from MONICA™ devices. NoSQL DB are usually faster for inserting data coming at a high frequency from sensors, on top of this the flexible schema of MongoDB is quite useful in contexts like this

In facts, data can be accessed through AQ maps relying on Vue.JS™ [28] a Javascript framework for interactive web interface definition and implementation, and Leaflet™, a web scripting framework for interactive maps visualization. The choice of VueJs allowed a full stack JavaScript workflow, allowing to shorten the development time and is a quite common solution in IoT scenarios. Finally, an Android based App have been developed to receive data via the BLE interface during operative deployment. It is also tasked for geolocation and remote data transmission towards the inception layer [9]. Designed to host a calibration implementation class, it is capable to locally return an immediate feedback to the user allowing both personal exposure monitoring and the transmission of concentration readings avoiding now undue raw sensors data while allowing geostatistic based multiple sensors data fusion on the cloud. Complete schema is reported in Supplemental Materials. Obviously, as every solution, it has pros and cons, and it is certainly not the only possible choice, as an example *InfluxDB™* represent a valid alternative to MongoDB allowing for faster query execution in interactive data analysis scenarios.

## IV. METHODOLOGY

### A. Dataset composition

The devices have been collocated three times in a 1.5 yrs period (Winter 2020/2021, Summer 2021, Winter 2021/2022) in order to derive calibrations functions in view of 3 different, citizen driven, participatory monitoring campaigns whose results have only been partially published. During winter 2020/2021 colocation experiment (*from Jan, 13th 15:00 to Mar 24th 10:00*), the 30 devices have been partitioned in 3 equal sized sets of 10 devices each. Each set have been continuously collocated for 3 weeks on the roof of the reference station located at less than 20m from a main street in the city of Portici, 7 km south of Naples city centre in the south of Italy (40°49'18.3"N 14°19'27.8"E), determining a 3 periods partition. Sampled data have been averaged at 1 hr rate and synchronized with reference analyzer data stream to form a dataset used for calibration and validation purposes. A similar colocation experiment has then been repeated during the summer/autumn time of the same year (*from Jul, 4th 00:00 to Oct, 4th 9:20*) resulting in a further dataset used for replication and long term validation purposes (see Table 1 for timing details). Some of the device failed to complete the full deployment terms due to maintenance needs. As such, the dataset shows variance in the device availability across different deployment periods and this may affect comparability of absolute performance indicator values across these periods (e.g. winter, summer), however the comparability across the two presented different calibration methodologies is basically unaffected.

---

[1] Campania Regional Environmental Protection Agency

## B. Multi-Unit Field Calibration Methodology

During the colocations, both concentrations of target particulate fractions and humidity levels changed significantly. Generally, lower concentrations and humidity levels were observed during the summer deployment (see Fig. 3). During Winter time, actually, the second period/batch was characterized by highest recorded pollution levels while the lowest were recorded during first period of the summer deployment. These frequently and widely observed variabilities are known to induce loss of accuracy in field calibrations when used under regimes which are different from the ones in which they have been derived [10]. This issue challenges the robustness of such strategy which is already negatively affected by the mentioned scalability problems. Our goal here was **to test, under different pollution and environmental regimes, the efficiency and accuracy of a multi-unit calibration obtained by the use of data samples recorded by a small number of devices (calibration devices subset)** while being simultaneously collocated with reference analyzers. More precisely, we proposed and tested a global, universal calibration law, generally applicable for all the devices of the same model, **comparing it with traditional, low scalability process** which aims at deriving a specific (ad-hoc) calibration for each different device.

TABLE 1: COLOCATION TIMING, 2021 WINTER AND SUMMER CAMPAIGNS.

| Deployment | Period 1 | Period 2 | Period 3 |
|---|---|---|---|
| Winter | Jan, 13th 15:00 -> Feb, 5 12:00; | Feb, 5th 12:00 -> Mar 2nd, 10:00; | Mar, 2nd 14:00 -> Mar 24th 10:00 |
| Summer | Jul, 4th 00:00 -> Jul 19th 23:59; | Aug, 24th 11:00 -> Sep, 14th 8:40; | Sep 14th 10:15 -> Oct, 4th 9:20 |

The proposed global calibration approach can be exemplified by the following steps (fig. 2):

1. Collecting raw sensor data from a small number of LCAQMS, same type units, when collocated with a reference analyzer to provide for true concentrations values
2. Fusing the collected different devices dataset into a single raw database
3. Use raw sensor data along with the correspondent reference data to derive the optimal parameters for the calibration law

We then apply the derived calibration law to all other same type units comparing it with ad-hoc derived calibration (fig.4)

We used two data fusion methodologies for the data in the calibration devices subset:

- **a-Aggregation:** raw data coming from different devices are separately averaged at hourly rates and then aggregated in the same dataset resulted in $n$ x $size$ points dataset where $n$ is the number of calibration devices and $size$ is the number of hrs corresponding to calibration set duration.
- **b-Median:** raw data collected from different devices are averaged at hourly rates and then subjected to median computation feature-wise and the result is stored into the median response dataset with size equal to the number of hours corresponding to the calibration set duration.

While *aggregation* provides a very simple data fusing method at a very low computational cost, *median* method allows for reducing the size of the training dataset and hence the computational cost of the training phase. However, the nodes must also be collocated in the same timeframe to derive the median while in this regard, *aggregation* provides for more flexibility. At the core of both methodologies, namely ad-hoc and global, lies the following basic multilinear calibration law:

$$C_{PM_x}(t) = aPM'_x(t) + bRH(t) + c \qquad (1)$$

with $PM'_x$ being vendor estimate for size $x$ in [2.5,10] um fractionated PM concentration and $RH$ the MONICA™ sensor estimate for relative humidity. This law hence translates raw vendor based concentration estimations, corrected for RH interference, into accurate targeted size particulate concentrations in real time. This basic algorithmic structure is aiming to a limited computational impact on the duty cycle of the executing device whether being the very same LCAQMS device, a companion device (such as a smartphone) or an edge appliance. In order to test for the accuracy of the proposed methodology we devised 2 separate experiments targeted to capture *short* and *long term* behavior of the devised approach, respectively. For the *short-term* behavior, we start by considering 2021/2022 winter time dataset. This dataset have been partitioned in **3 different continuous periods/batches** each one containing data related with 10 devices separately. In turn, 10 devices and so one of the 3 colocation periods are chosen to take part in the global calibration devices subset while the others, collocated in the remaining two periods, take part in the validation devices subset (fig. 5). For each of the 3 possible choices of the calibration device set, 2 out of the 3 weeks of the calibration devices set colocation period is used as a training set for machine learning (ML) based multi-unit (global) calibration approach. In this specific case as detailed in (eq. 1) we rely on Multi-Linear Regression (MLR). Data are fused according to method *a* or *b* as described above.

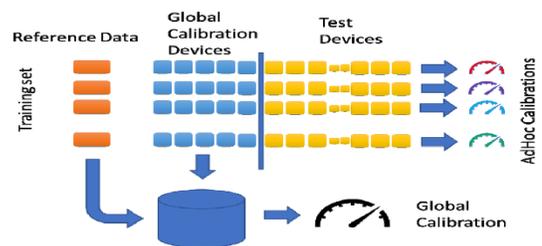

Figure 2: In the proposed approach, a limited number of devices is used to derive a single multi-unit (global/universal) calibration through data fusion and machine learning approach. The performance of the obtained model is compared with the performance obtainable by AdHoc calibration method

For each choice of the global calibration subset, 2 out of three weeks of the colocation time of each of devices belonging to the remaining validation device subset are chosen as training set for deriving the parameters of the same device ML calibration model (see supplemental materials, flowchart 1).

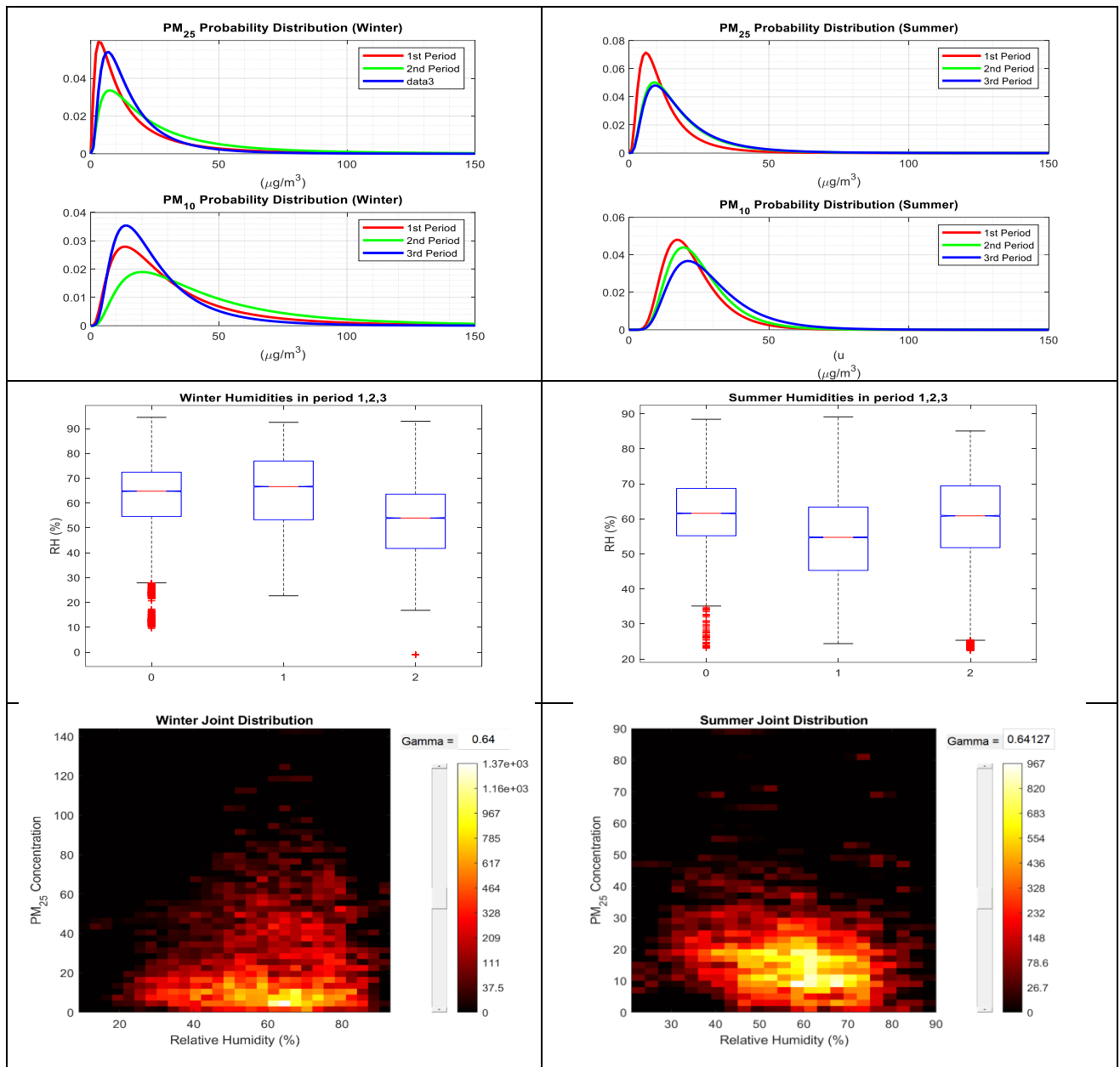

Figure 3: Probability distribution under Lognorm distribution hypothesis of PM fractions concentrations during the 3 colocation periods in Winter (a) and Summer (b). Box Plot of Relative Humidity recordings during Winter (c) and Summer (d) colocation periods. Joint Histogram (concentrations in ug/m3) distribution of both forcers during whole Winter (e) and Summer (f) colocation periods.

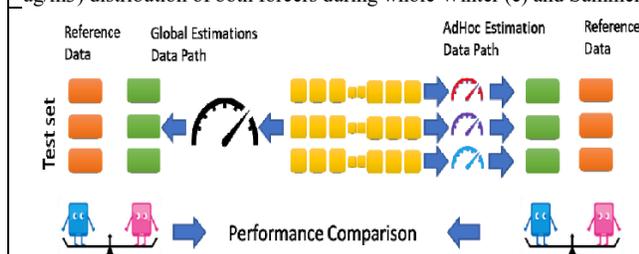

Figure 4: Performance comparison process. For each test device and for each relevant time instant (1 hour), a unified global calibration model is tested on the same data used to assess the performance of ad-hoc models, i.e. one derived for that specific device. Regulatory grade analyzers provide for true concentrations used for performance rating computations. Performance is computed for each device test set and averaged across all tested devices before being compared.

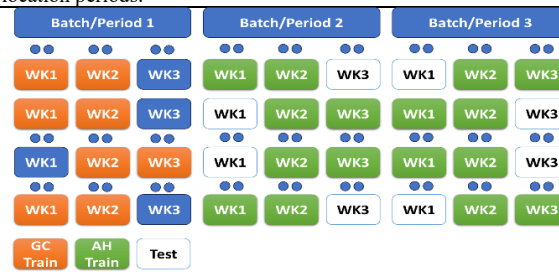

Figure 5: Examples of combinations of Global Training set week composition and corresponding AdHoc Training set compositions. Once the global training batch have been chosen along with the participating devices, the obtained calibration is tested across all possible test set week and compared with the corresponding results of the AdHoc algorithm.

As a remark, in this case data are not fused across different devices but data coming from a single device was used to derive an *ad-hoc calibration* law to be used just for that specific device.

In the end, for each of the devices belonging to the validation dataset, the global calibration law and the specifically derived ad-hoc calibration are applied during the remaining one week of colocation which is the final test set. Performances are computed and results are averaged across all possible 3 choices of the inner test set period (1 week) for all the 20 test devices. The entire process is then repeated for all the possible choices of the 3 device batch set and the partial, single calibration batch dependent, results (see figures 6 and 7) are averaged again to provide the final assessment of the performance of both global and ad-hoc procedures[1]. In facts, all the possible choices of 10 calibration devices set, and hence of the training weeks as well as of the test weeks are considered so to limit the dependance of the performance assessment from the specific calibration devices batch and from the peculiar joint distribution of the particulate concentrations and environmental forces occurred during all the possible combinations of training and test periods. It's interesting to note that given the slowly changing environmental and emission conditions, in this comparison scenario, the global calibration approach is slightly penalized by the increased time distance between training and test set with respect to ad-hoc calibration models. The latter are derived in the same 3 weeks continuous colocation time slot as the test week. By contrast the global calibration approach, in its aggregated fashion, may numerically benefit from the enhanced number of samples though being strongly correlated. It's equally important to note that once the batch have been chosen, the 10 devices are sorted in a specific random order. Then the performances are assessed using the first $k$ devices chosen from the sorted list to take part in the calibration device dataset, $k$ obviously range from 1 to 10. This is done aiming to study the relationship between performance and the number of used devices. The resulting accuracy indicator curve shape depends on the specific order dictating the specific combination of devices used in the sets of 1 up to 10 devices. To eliminate the dependency on the chosen order, the entire sequence choice process is repeated 100 times averaging the performances, across the different k-long generated sequences which guarantees a uniform sampling of the devices combinations space. The summer time deployment was partially different in the number of collocated devices for each of the 3 periods. Actually the first periods saw the deployment 8 devices, while the second included 5 devices. The final deployment included 14 devices. Practically, the performance comparison among the global and ad-hoc calibration approaches described above could only be conducted for a global calibration device subset of up to 5 devices. In principle it is possible to compare the results obtained by using up to 14 devices, but of course starting from the sixth device on we are only averaging across 2 periods while starting from 9 devices on, we would consider only one single period of deployment with larger expected uncertainties. Results obtained during the summer time are reported in the supplemental materials.

For assessing *long-term* behavior, the testing process is somewhat different. Here the models were derived using complete winter datasets, and then, in order to test long term behaviour, tested in the complete summer period. In more details, this was done in the following way. Winter dataset contained 3 batches of MONICA devices (see Table S.1 for more detailed description of the datasets and the timeline of calibration). Calibration models were computed for each batch using data from 1, up to 10, the maximum number of MONICA devices in each batch of the dataset. All possible units combinations were used to derive the models with a brute-force approach. As an example, if the batch contains 10 MONICA devices, there are 10 calibration models that were trained on 1 device, $\binom{10}{2} = 45$ models that were trained on 2 devices, $\binom{10}{3} = 120$ models that were trained on 3 devices, etc. To test the long term behaviour of the models that were trained in this way using only winter data, each of the models was tested in the summer period. The testing data points come from all possible combinations of individual models and MONICA device data from the summer period that are eligible for testing. Here we use more strict criterions than what is typically used in machine learning practice, since we want to test both long term behaviour (thus winter training, summer testing separation) and transferability of the models (MONICA devices used for testing must not come from the batch of MONICAs used for development of the model). Ad hoc calibration models are tested without transfer of calibration and with only one device used for model training, i.e. same MONICA device was used for training (winter time) and testing (summer time).In both experiment we consider Mean Absolute Error (MAE) **MAE** and **R²** to compare the performance of the different approaches. They are among the most commonly used performance indicators used for low cost air quality sensor performance assessment. MAE is computed through the following formula:

$$MAE = \sum_{t=1}^{T} |C_{PM_x}(t) - C^*_{PM_x}(t)|$$

With $C^*_{PM_x}(t)$ being true concentration level at the specific time instant and provides for an quantitative assessment of the average of the absolute difference between estimated and true concentration and its sensitivity to outliers, which occurs frequently for LCAQMS, is less pronounced with respect to RMSE. However it has the evident shortcoming of masking the relative magnitude of the error at different concentration levels for which it is generally coupled with **R²** providing a goodness-of-fit quantitative estimation.

## V. Experimental Results

### A. Short-term Experiment

#### i. Experimental Implementation

A preliminary evaluation of average performances of vendor based calibration for PM$_{2.5}$ was performed resulting in values listed in table 2 depicting PM2.5 Average (std) of performance indicators values. obtained by vendor based calibration on the same short-term evaluation setting of ad-hoc performances (30 sensors x 3 inner testing periods = 90 total evaluations, for winter time; 27 x 3 =81 total evaluations for summer time).

---

[1] Computed performance indicators populations are stored for uncertainty analysis and reporting (see figures 6 -13) and statistical tests implementations.

TABLE 2: PM2.5 AVERAGE (STD) OF PERFORMANCE INDICATORS VALUES.

|  | MAE | R^2 | RMSE | NRMSE | MAE/RANGE |
|---|---|---|---|---|---|
| **Winter 2021** | 10.1 (4.69) | 0.21 (0.44) | 13.59 (5.48) | 0.86 (0.25) | 0.13 (0.06) |
| **Summer 2021** | 8.7 (2.3) | 0.25 (0.25) | 12.39 (3.4) | 0.86 (0.14) | 0.10 (0.03) |

Those values have been obtained exactly in the same experimental conditions of short-term experiment described above. As such they can be compared with ad-hoc calibration performance evaluation in short-term experiment. Afterwise, procedures described in chapter 5, relatively to short-term behaviour assesment, have been implemented in Matlab™ and run on a multicore PC (Intel(R) Core(TM) i7-10750H CPU @ 2.60GHz, 16GB RAM, NVIDIA GTX 1650TI GPU). Collected results have been graphically rendered in figure 6 to 13, depicting average $R^2$ and MAE performance indicators as a function of the number of devices involved in the global calibration set for both the aggregated and median data fusion techniques for PM2.5 (respectively figs. 6-10) and PM10 (figs. 11-13). In both cases, performance indicators have been compared with the respective average figures obtained by ad hoc calibrations on the same devices test sets.

### ii. Performance comparison approach

Variance bands in Figure 8 and 11 (a to d) depict observed variability in sampled averaged performance indicators with a 1-σ large interval around sample average reflecting the variability in performance induced by the choice of the global calibration training period (and consequently of the resulting two test periods) as well as the associated global calibration devices set (and respectively the devices test sets). For the global calibration side only, it also take into account the variance induced by the different combinations of n devices that actually may have taken part in the calibration sets in the 100 different extractions when their size n goes from 1 to 10. Similarly, confidence bands in Figure 9 and 12 (a to d) depict .95 confidence level (alpha=0.05) in the determination of performance indicator sample average. Confidence intervals are specifically drawn to help to evaluate the performance that ad-hoc and global method may achieve in the general case. In this case, for uncertainty computations and analysis, we have considered a model characterized by two uncertainty components: the first one arise from the choice of the training deployment period (which also determine the training device population and test devices sets), the second one arising from the choice of the composition of the global training device subset. So we can describe the measurements of the specific performance index $\delta_i$ as:

$$\delta_i = \theta + \varepsilon_i$$

where $\theta$ is the value of the performance index on a generic test set and $\varepsilon_i$ an error term accounting for the variance induced by the specific composition of the global calibration device set. According to this model's uncertainty, due to the inherent correlation in performance indices values by global calibration approaches in the same test period, uncertainty is determined by summing the uncertainty on the expected value of the index across different deployments periods (degree of freedom=3) with the uncertainty on the expected value of the index across different composition of the global training device subset (degree of freedom = min (100, $\binom{n}{k}$)) with n being the total number of available global training period devices and k being the number of involved global calibration devices. Here, it is important to note that while the ad hoc performance indicator average, computed over 20 test devices, can only take 3 values depending on the three different possible compositions of the device test set, the average performance of the global calibration algorithm on 20 devices may here take up to 3 x 100 values. Finally figure 10 and 13 (a and b) directly compares the results obtained by the two proposed data fusion schemes i.e. response aggregation and response median computation using confidence intervals.

Observed difference have been tested for statistical significance and results have been reported in supplemental materials table 2 to 4. Wilcoxon signed rank (non parametric) test have been used to formally evaluate the hypothesis of paired observed differences between the curves at each n to belong to a zero median distributed population. Furthermore paired t-Test is used for testing the null hypothesis of the two populations belonging to equal mean and different variance distributions except for global calibration methodologies direct comparison. In the paired t-test case, care should be put on the evaluation of the results because Jarque-Bera test [29] shown that no performance indicator average population at any n in any experiment have shown to be sampled from a normal population nor their depicted difference. Furthermore the two compared population are always related since the average performance indicator is computed on the same device set on same sensor data. However at this population size, it is usually accepted to be tested with t-test class tests in view of the central limit theorem.

### iii. Short-term Results analysis

Results shows that, in general terms all the approaches reach satisfying performance in the short-term both in terms of accuracy. When comparing global calibration approaches and standard ad-hoc ones, while $R^2$ at some n for both aggregation and median method seems to indicate slight or no statistically significant difference in the performances, MAE results always indicates a slight but statistically significant advantage of the global calibration methods in terms of non zero median in the recorded difference. All tests have been conducted at alpha = 0.05 significance level. It is also worth to note that fig.5 and 6 shows that variance in the recorded performance of global methods is largely dominated by the different scores obtained in the different 3 test periods while being negatively correlated with the number of involved devices as long as the number of degree of freedom i.e. the number of available combinations of units is sufficiently large. Hence, we can conclude that on short-term deployment scenarios, global (multi-unit) calibration models shows on par or slightly better performances while dramatically improving the scalability.

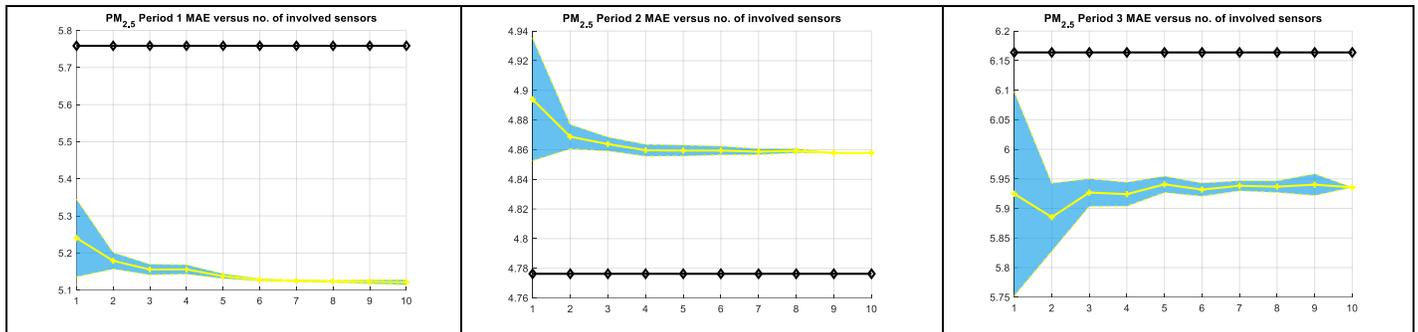

Figure 6: MAE (ug/m3) of Global Calibration Estimations obtained on average on period 1,2,3 respectively (a,b,c). Global calibration have shown a positive edge on ad-hoc calibration in 2 out of 3 testing periods., Confidence intervals show a decreasing variance with the increase of the number of involved devices.

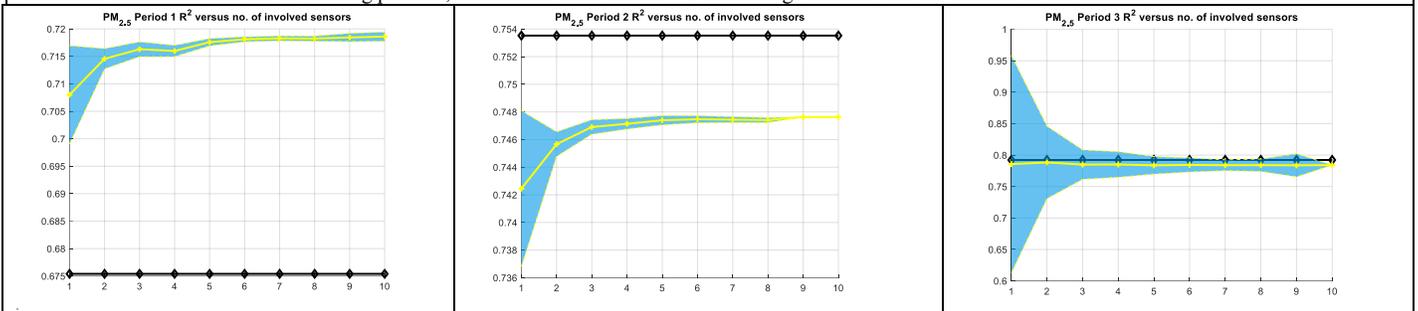

Figure7: $R^2$ of Global Calibration Estimations obtained on average on Winter period 1,2,3 respectively (a,b,c). Global calibration have shown better performances with respect to ad-hoc calibration in 1 out of 3 testing periods, on par performance on 1 period and worse performance in the remaining one. Confidence intervals show a decreasing variance with the increase of the number of involved devices.

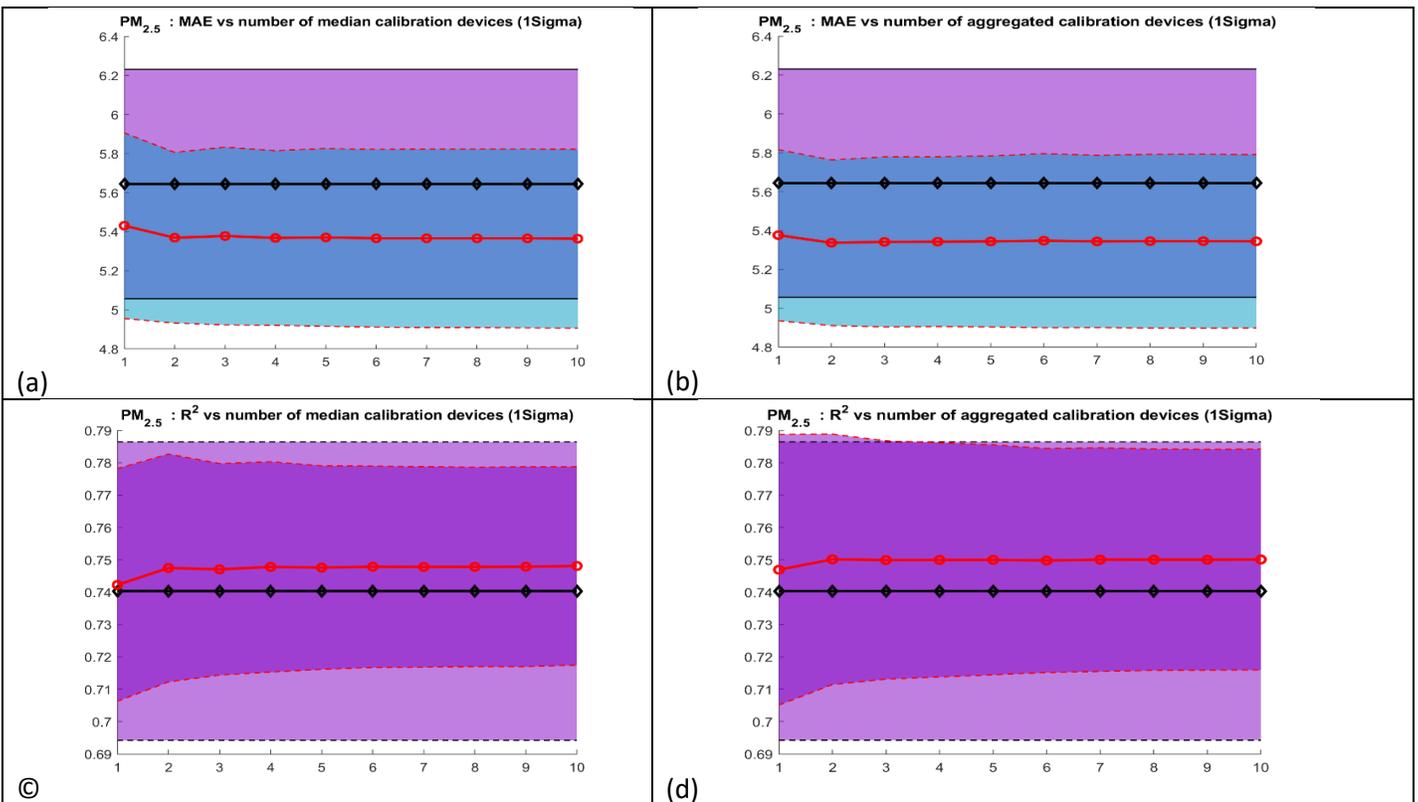

Figure 8: Winter Colocation Global Calibration approaches (red) versus Ad-Hoc Calibration (black) average MAE (a,b) and R^2 (c,d) figures at different no. (n) of involved devices along with variability bars (1-sigma). Performance indices evaluation are affected by variance induced by the choice of both test period and composition of calibration devices set, however test period induced variance is more relevant.

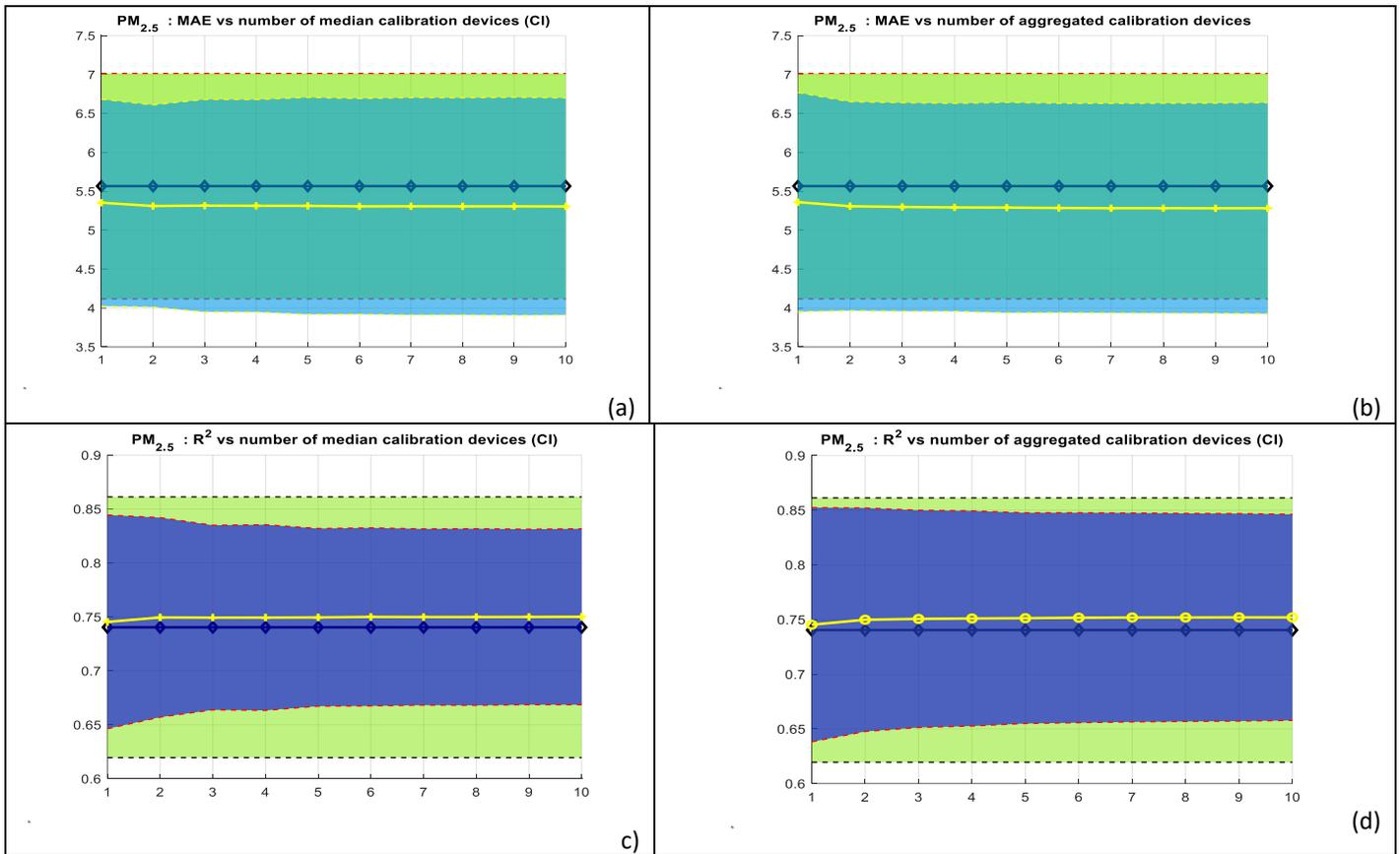

Figure 9: Winter Colocation Global Calibration approaches (yellow) versus Ad-Hoc Calibration (black) average MAE **(a,b)** and R^2 **(c,d)** figures at different no. **(n)** of involved devices along with uncertainty bars (CI).

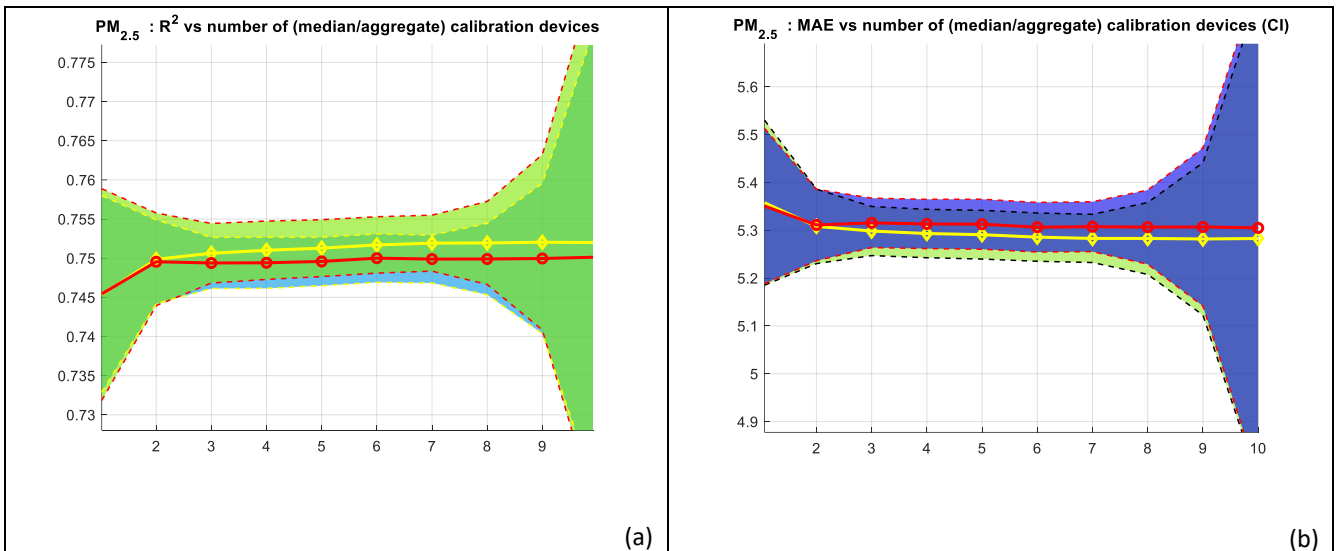

Figure 10: Winter Colocation Comparison of the two data fusion models (red: median; yellow:aggregate) for the proposed global calibration methodology using uncertainty bars (0.95 CI). While aggregated models consistently but very slightly outperform median based models, uncertainty bands of both performance index population means are actually largely superimposed.

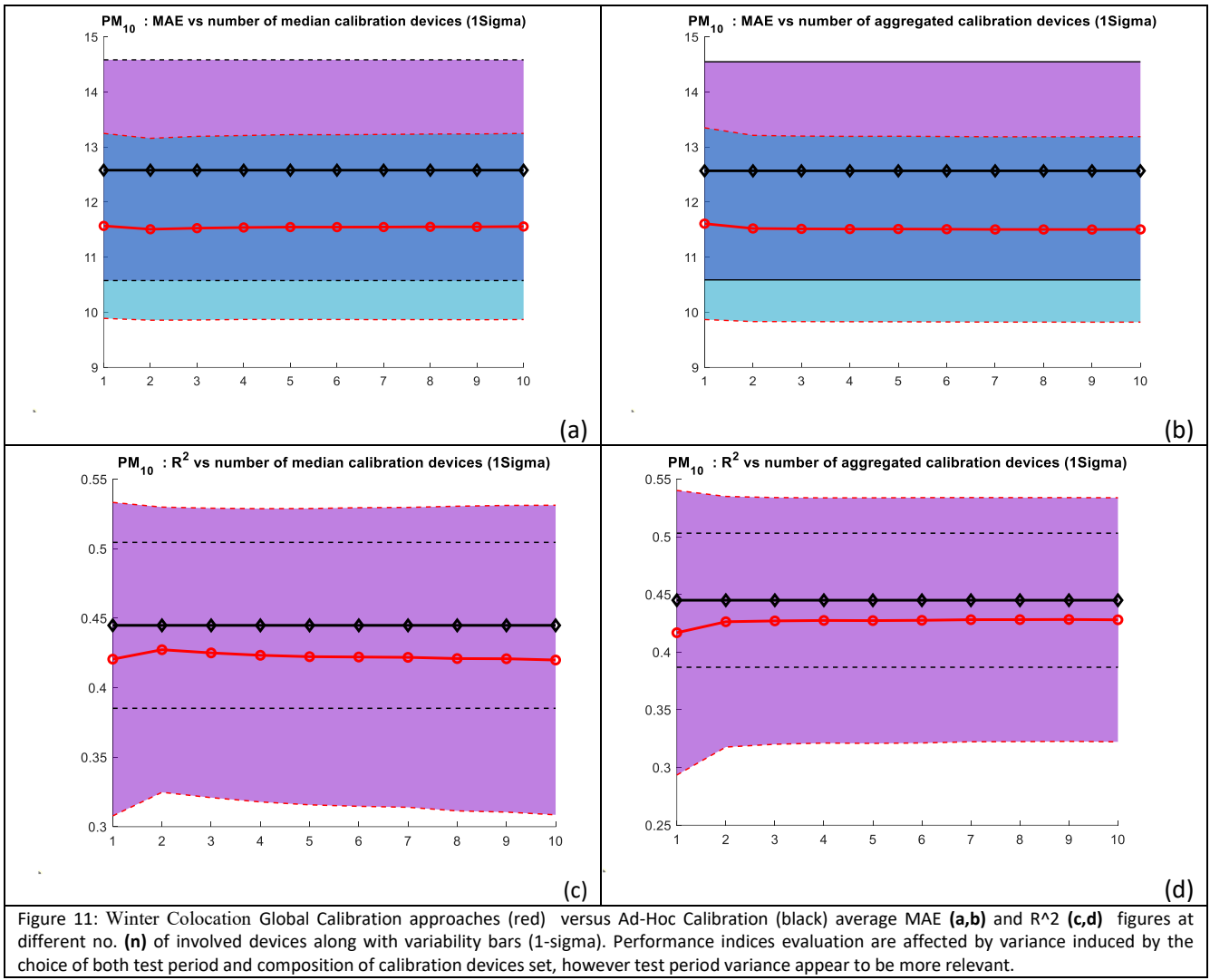

Figure 11: Winter Colocation Global Calibration approaches (red) versus Ad-Hoc Calibration (black) average MAE **(a,b)** and R^2 **(c,d)** figures at different no. **(n)** of involved devices along with variability bars (1-sigma). Performance indices evaluation are affected by variance induced by the choice of both test period and composition of calibration devices set, however test period variance appear to be more relevant.

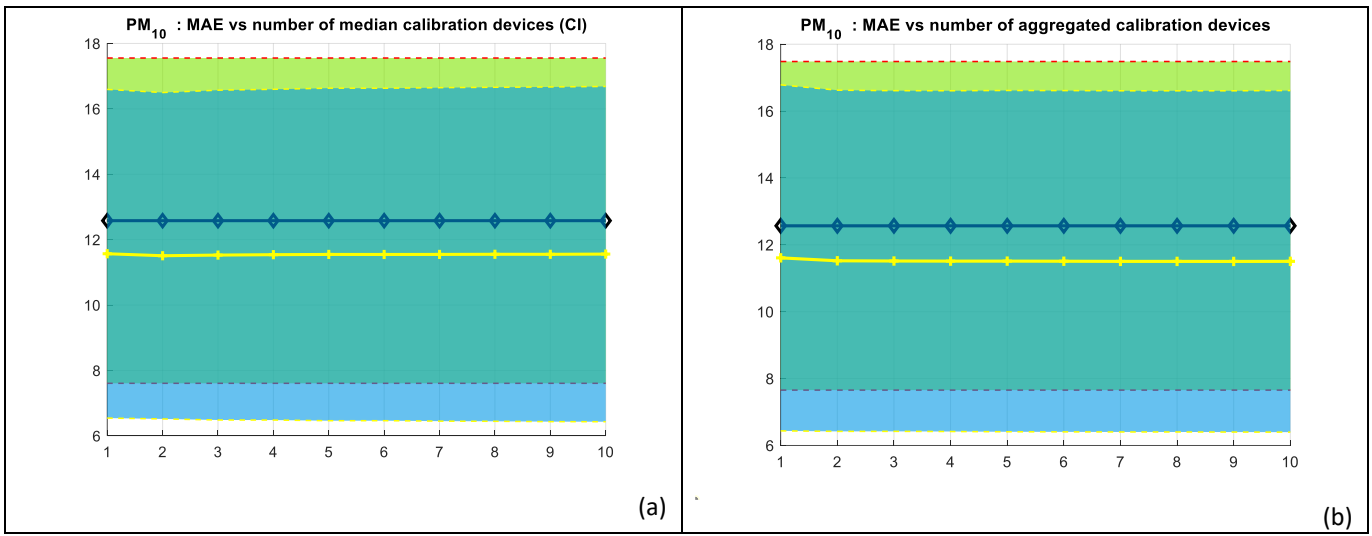

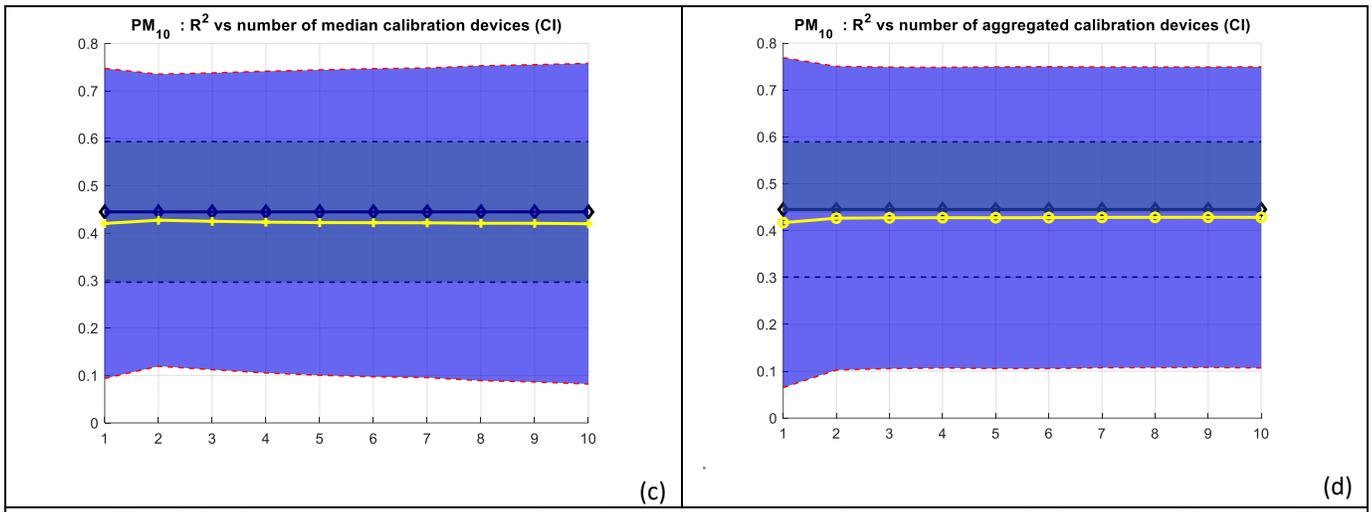

Figure 12: Winter Colocation Global Calibration approaches (yellow) versus Ad-Hoc Calibration (black) average MAE (a,b) and R^2 (c,d) figures at different no. (n) of involved devices along with uncertainty bars (CI). The figures graphically explains the results of statistical significance tests among the sample population of R^2 and MAE when average across the test devices and periods.

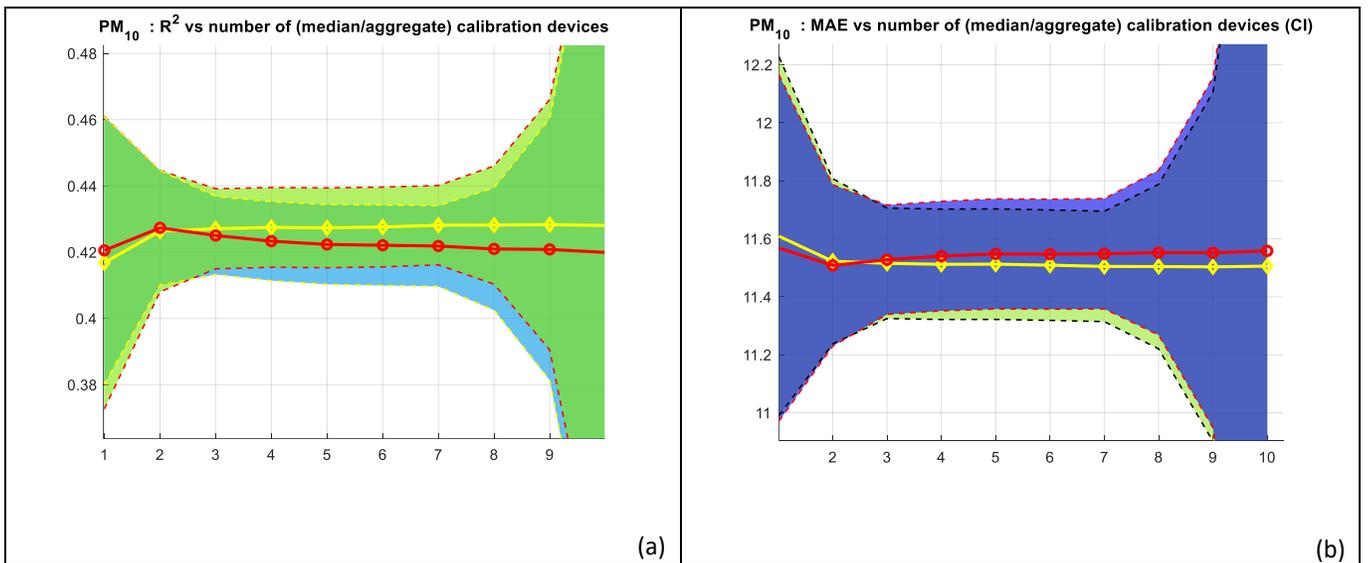

Figure 13: Winter Colocation Comparison of the two data fusion models (red: median; yellow:aggregate) for the proposed global calibration methodology using uncertainty bars (0.95 CI). While aggregated models consistently but very slightly outperform median based models, uncertainty bands of both performance index population means are actually largely superimposed.

Conversely formal tests between aggregated and median methods have shown no statistical significance in the observed differences although showing a faint but consistent advantage of aggregate method over median method over all calibration device set size (1 to 10).

It is interesting to note that, differently from the median method, aggregate method do not require simultaneous readings from all the sensors for a specific time data point to be included in the training dataset.

Similar results in terms of comparison between global and ad-hoc approaches have been obtained by looking at PM10 concentration estimation problem. General performance however are sensibly lower than those obtained by both calibration approach with the 2.5 micron fraction. Results are shown in figure 10 to 12. Summer time results are instead reported in the Supplemental Materials section.

### A. Long Term Experiment

#### a. Experimental Implementation and Performance comparison approach

For long term performance analysis, we focus on PM2.5 results and aggregation based data fusion, assuming the generalization of the results given the previous findings. Calibrations obtained using winter data with aggregated multiunits dataset from a single deployment batch, have been tested on summer time data for set apart units.

Note that for long term, we have combinatorially scanned the complete set of possible attainable global calibration models for each batch, thus boxplots in figures below show complete bounds of performances for global calibration in long term scenario. Figures 13 and 14 have been designed to show long term performance for PM2.5 calibration models with intercept and 2 predictors (PM2.5 low cost sensor readout and relative humidity), developed using aggregated data from 1 to 10 calibration devices. Table 3 and 4 have been used for comparing R2 and MAE performance indicators quantiles across the 3 different batches for global calibration while ad-hoc results are reported in Table 5. Table 6 provides for a comparison view.

#### b. Long term results analysis

Fig. 14 and 15 compare performances as captured by MAE and $R^2$ indicators while varying the number of devices involved in global calibration derivation.

Table 3 to 5 show interquartile ranges and medians of R2 and MAE performance metrics of the global and ad-hoc models while a comparison between the two approaches is quantitatively given in table 6. Based on the reported data, it is clear that the variations in performance among the global calibration models, derived using different numbers of aggregated calibration devices, are very subtle (refer to Tables 3 and 4).

Moreover, performance seems to stabilize significantly when using five or more units. There is a notable reduction in both interquartile and variance intervals in MAE figures when increasing the number of involved devices. Figure 14 also illustrates a slight decrease in the overall spread of R2 under the same conditions. However, the median performance for both indicators remains quite similar. The transferability of the global calibration to test devices appears to be favorable, given that its results align well with the ad-hoc model performance. The fair assessment of the median $R^2$. which lies within the range of 0.3, should be conducted considering the inherently challenging task of ensuring long-term robustness in a field-derived calibration.

When transferred and tested, we observed that a few models exhibit negative R2 values, indicating that these models perform worse than using a constant value. However, such variance is also present in the ad-hoc models, underlining the challenge of maintaining accuracy in the summer due to changing operating conditions when additionally the concentration values are close to the sensors' limit of detection. The relatively low MAE values are also influenced by the relatively low concentrations of pollutants recorded during the summer months.

We remind here that ad-hoc is the best achievable approach which is, on the other hand, very difficult to implement in practice due to complicated logistics. In table 6, performance metrics quantiles, averaged across all 3 batches (for 5+ devices) are compared with adHoc calibration obtaining a faint edge over Global calibration while this appears reversed for MAE. To all practical means, the two approaches appear equivalent. Summarizing, global calibration performs on par with the ad-hoc model also in the long term.

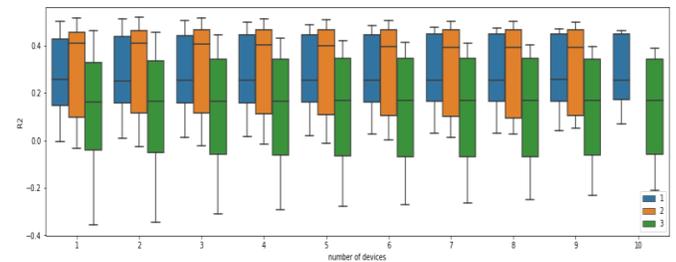

Figure 14: PM2.5 long term experiment: $R^2$ boxplot vs number of aggregated calibration devices for global calibration models (blue-Batch 1, orange-Batch 2, green- Batch 3)

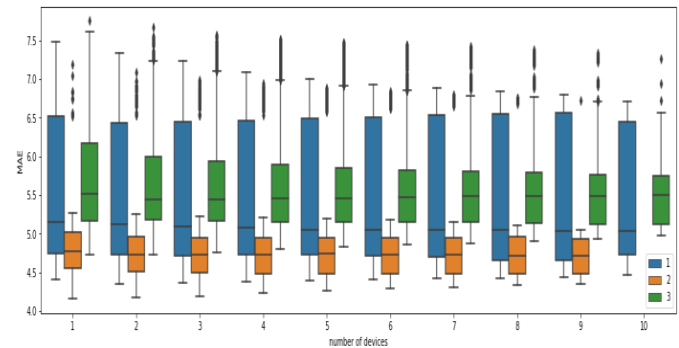

Figure 15: PM2.5 long term experiment: MAE boxplot vs number of aggregated calibration devices for global calibration models (blue-Batch 1, orange-Batch 2, green- Batch 3)

TABLE 3: QUANTILES (25%, 50%, 75%) OF $R^2$ PERFORMANCE METRICS FOR BATCHES 1,2 AND 3 FOR GLOBAL CALIBRATION MODELS

| | Batch 1 | | | Batch 2 | | | Batch 3 | | |
|---|---|---|---|---|---|---|---|---|---|
| | *Quant.* | | | *Quant.* | | | *Quant.* | | |
| *Dev.* | *0.25* | *0.5* | *0.75* | *0.25* | *0.5* | *0.75* | *0.25* | *0.5* | *0.75* |
| 1 | 0.15 | 0.26 | 0.43 | 0.09 | **0.41** | 0.46 | -0.04 | 0.16 | 0.33 |
| 2 | 0.16 | 0.25 | 0.44 | 0.12 | 0.40 | 0.46 | -0.05 | 0.16 | 0.34 |
| 3 | 0.15 | 0.25 | 0.44 | 0.11 | 0.40 | 0.46 | -0.05 | 0.16 | 0.34 |
| 4 | 0.16 | 0.25 | 0.44 | 0.11 | 0.40 | 0.47 | -0.06 | **0.17** | 0.34 |
| 5 | 0.17 | **0.26** | 0.44 | 0.10 | 0.40 | 0.47 | -0.06 | 0.17 | 0.34 |
| 6 | 0.16 | 0.25 | 0.45 | 0.10 | 0.40 | 0.47 | -0.07 | 0.17 | 0.34 |
| 7 | 0.16 | 0.25 | 0.44 | 0.09 | 0.39 | 0.46 | -0.06 | 0.16 | 0.34 |
| 8 | 0.16 | 0.25 | 0.44 | 0.09 | 0.39 | 0.46 | -0.06 | 0.16 | 0.34 |
| 9 | 0.16 | 0.25 | 0.44 | 0.10 | 0.39 | 0.46 | -0.06 | 0.16 | 0.34 |
| 10 | 0.17 | 0.25 | 0.44 | nan | nan | nan | -0.05 | 0.16 | 0.34 |

TABLE 4: QUANTILES (25%, 50%, 75%) OF MAE PERFORMANCE METRICS FOR BATCHES 1,2 AND 3 FOR GLOBAL CALIBRATION MODELS

| | Batch 1 | | | Batch 2 | | | Batch 3 | | |
|---|---|---|---|---|---|---|---|---|---|
| | *Quant.* | | | *Quant.* | | | *Quant.* | | |
| *Dev.* | *0.25* | *0.5* | *0.75* | *0.25* | *0.5* | *0.75* | *0.25* | *0.5* | *0.75* |
| 1 | 4.73 | 5.15 | 6.52 | 4.5 | 4.76 | 5.01 | 5.16 | 5.51 | 6.17 |
| 2 | 4.72 | 5.12 | 6.42 | 4.5 | 4.72 | 4.96 | 5.17 | **5.43** | 6.00 |
| 3 | 4.71 | 5.09 | 6.44 | 4.4 | 4.72 | 4.95 | 5.16 | 5.44 | 5.93 |
| 4 | 4.72 | 5.07 | 6.46 | 4.4 | 4.72 | 4.94 | 5.15 | 5.45 | 5.88 |
| 5 | 4.72 | 5.05 | 6.48 | 4.4 | 4.74 | 4.94 | 5.14 | 5.46 | 5.85 |
| 6 | 4.70 | 5.05 | 6.51 | 4.4 | 4.73 | 4.94 | 5.14 | 5.46 | 5.82 |
| 7 | 4.69 | 5.04 | 6.52 | 4.4 | 4.72 | 4.95 | 5.14 | 5.48 | 5.81 |
| 8 | 4.65 | 5.04 | 6.54 | 4.4 | 4.71 | 4.95 | 5.13 | 5.48 | 5.79 |
| 9 | 4.65 | 5.04 | 6.56 | 4.4 | **4.70** | 4.92 | 5.12 | 5.49 | 5.75 |
| 10 | 4.73 | **5.03** | 6.44 | nan | nan | nan | 5.12 | 5.50 | 5.75 |

TABLE 5: QUANTILES (25%, 50%, 75%) OF $R^2$ AND MAE PERFORMANCE METRICS FOR BATCHES 1,2 AND 3 FOR AD-HOC CALIBRATION MODELS

| Batch 1 | | | Batch 2 | | | Batch 3 | | |
|---|---|---|---|---|---|---|---|---|
| *Quantiles* | | | *Quantiles* | | | *Quantiles* | | |
| *0.25* | *0.5* | *0.75* | *0.25* | *0.5* | *0.75* | *0.25* | *0.5* | *0.75* |
| **$R^2$ Quantiles** | | | **$R^2$ Quantiles** | | | **$R^2$ Quantiles** | | |
| 0.05 | **0.30** | 0.47 | 0.20 | **0.26** | 0.46 | 0.04 | **0.29** | 0.35 |
| **MAE Quantiles** | | | **MAE Quantiles** | | | **MAE Quantiles** | | |
| 4.52 | **4.57** | 4.90 | 4.78 | **5.59** | 6.39 | 5.18 | **5.43** | 5.72 |

TABLE 6: COMPARISON BETWEEN GLOBAL AND ADHOC CALIBRATION METHODOLOGIES.

| **All Batches Quantiles** | | |
|---|---|---|
| *0.25* | *0.5* | *0.75* |
| **$R^2$** | | |
| Global | 0.067 | 0.27 | 0.42 |
| **adHoc** | **0.10** | **0.28** | **0.43** |
| **MAE** | | |
| **Global** | 4.77 | **5.08** | 5.75 |
| adHoc | 4.83 | 5.20 | **5.67** |

## VI. DISCUSSION AND CONCLUSIONS

This study exploited the data recorded during a multi-seasonal colocation of Low cost Air Quality Multisensor devices to compare 0-sample transferable multiunit calibration strategies, derived with data recorded from a limited device subset, with state of the art ad-hoc field calibration strategy. This, being still portrayed in the literature as the one delivering the best performance, have shown clear scalability limits which actually prevent large scale commercial deployments of these IoT device. The comparison has been conducted using two among the most recurrent performance indicators in literature aiming to capture both error and correlation aspects [22] and taking particular care to fairness of the comparison. Actually, multiunit strategies have been subjected to slightly disadvantaged conditions, i.e. operating its estimations in test periods located timewise farther from the calibration time, with respect to the ad-hoc strategy. Nonetheless, the analysis of the reported results allowed to state, with a significant level of confidence and *to all practical purposes*, that multiunit field calibration represent a viable and scalable methodology for low cost particulate matter monitoring devices, **actually matching the performance obtainable with the ad-hoc strategy**. This could open the path to scale up deployments to hundreds of this IoT AQ monitoring devices with a cheaper investments in terms of logistics and manpower due to a more efficient calibration strategy. In terms of absolute results, the obtained MAE and R2 during short-term experiments are below (respectively exceeds) the typical acceptability levels for short-term LCAQMS deployments for hourly estimations of PM2.5 (7 ug/m3, 0.7) [8,22]. The low computational requirement algorithm can be implemented on board or edge devices putting the high quality insights gained at the hand of the device user for mobile and potentially wearable deployments. By choosing an on board execution model, it allows the immediate reduction of data transfer needs (from 12 *double* sized values to 3 *float sized values* per sample when considering $PM_{1/2.5/10}$ concentrations). However, model update may become more difficult requiring firmware updates with respect to edge/cloud implementation. In edge (or cloud) execution, on the other hand, will come at the cost of short range (respectively long range) raw data transmission need increasing data transfer rate and consumption. Two data fusion strategies have been tested with the purpose of exploiting data from calibration units: timewise aggregation and timewise median extraction. The two obtained very similar results with one (aggregation) showing very slight advantages further allowing a greater flexibility for device colocation periods choice potentially opening the path to in network federated learning. Eventually, the proposed methodology performances will suffer from impacts of changing environmental and emission conditions over time due to weather and anthropogenic/natural particulate emissions seasonalities. Like in the case of ad-hoc methodology, the limited variance in the short-term training set will not allow the calibration algorithm to reach the desired generalization properties to deal with very different statistical distribution of inputs. Ageing or poisoning effects, depending on a specific sensor long term operating conditions, will also contribute to long term negative impacts. In facts, further works will include the exploration of this latter methodology in increased diversity scenarios with respect to location and environmental conditions along with exploitation of deep learning algorithms for improved performance and generalization.

## ACKNOWLEDGMENTS

This work have received funding from European Commission H2020 Research and innovation program under **VIDIS** Project GA no. 952433. Authors desire to thank **ARPA Campania** and specifically **Dr. P. D' Auria** and **Dr. G. Onorati** for the availability for reference analyzers during the colocation campaigns, **Mr. G. Loffredo, Ing. A. Del Giudice** and **Ing. F. Formisano** (ENEA) for supporting the data gathering campaigns, **Dr. E Esposito** (ENEA) for data quality checks, **Dr. Santiago Marco** (IBEC, Barcelona) and **Dr. Ettore Massera** (ENEA) for interesting discussions on the subject.

AUTHOR CONTRIBUTIONS:

"Conceptualization, SDV and GDE; Investigation SDV, GDE and MD; Methodology & algorithms, SDV. SF and MD; Data curation, GDE, SF, DK and DS; Validation, All authors; Supervision, SDV and MJ; Project administration, MJ and GDF; Funding acquisition, SDV, GDF and MJ. All authors contributed to the original draft preparation."

**Corresponding Author Short Bio**

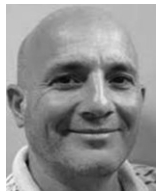

**Saverio De Vito** (M.S. '98, University of Naples "Federico II") received His Ph.D. degree in information engineering from the University of Cassino and South Lazio. From 1999 to 2004, he was an SW Architect specializing in TLC/EO platforms software. Since 2004, he has been a Full Researcher and a Team Leader with ENEA. His current research interests include artificial olfaction, IoT sensors, and Machine learning. His research has led to applications in environmental monitoring, energy production, aerospace industry, and water management. He coordinated/participated in several international research projects in FP7, H2020, FLAG-ERA, CLEANSKY, and UIA programs. He has co-authored more than 100 scientific contributions. He was a Professor of computer engineering with the University of Cassino, Cassino, Italy, from 2005 to 2020. Dr. De Vito is currently serving as President of ISOCS and is an ETIP-PV, and AISEM member. He serves for the UNI GL4 Group as well as IEEE P2520.1 standardization working groups. He is a TC Member and Guest Editor for several conferences in journals in measurements, computing, and environmental science journals.

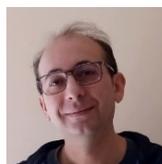

**Gerardo D'Elia** was born in Battipaglia, Italy, in 1977. He received the M.S. degree in electronic engineering and the Ph. D degree in industrial engineering from the University of Salerno, Italy, in 2004 and 2023, respectively.

From 2004 to 2013, he was with the Research and Development Department of a global telecommunications equipment company first as a Test Engineer and then as a New Product Industrialization Engineer. In 2014, he joined the Nuclear Fusion Department of ENEA, and in 2019, he joined the ENEA Energy Technologies and Renewable Sources Department in Portici, Italy, as a Research and Development Technician, where he was involved in air quality monitoring applying machine learning techniques (ML). Since 2022, he has been a Full Researcher at the same ENEA department. His current research interests include air quality monitoring technologies, wireless sensor network, Internet of Things (IoT), and ML techniques.

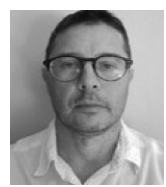

**Sergio Ferlito** was born in Naples, Italy, in 1969. He received the M.S. degree in electronic engineering from the University of Naples "Federico II," Naples, in 1998.

Since 2014, he has been a Full Researcher with ENEA, Naples, where he focuses in the field of forecasting applied to renewable. He has authored many research articles on photovoltaic power production forecasting and more generally on machine learning (ML) techniques applied to renewables. His current research interests include statistical and ML models applied to renewables and Internet of Things (IoT) fields, Web software development (front-end and back-end), and online ML frameworks. He is an Expert Technical Scuba Diver and holds some diving certifications.

Mr. Ferlito received a scholarship first and then received a research grant from ENEA, Italian National Agency for New Technologies, Energy and Sustainable Economic Development, in 1999, to work in the field of renewable, mainly photovoltaic, till 2012.

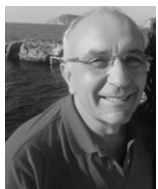

**Girolamo Di Francia** is a Research Director with ENEA, the Italian National Agency for New Technologies, Energy and Sustainable Economic Development, Naples, Italy. He is a Full Professor of applied physics, an Independent Expert-Supervisor for the European Union and for the Italian Minister for Research and Education in the areas of materials, new technologies, and renewable energies. During his scientific activity, he has authored or coauthored more than 220 papers in international journals (SCOPUS H-index = 29) and presented more than 250 communications to national and international congresses. He also holds 12 patents on solid-state electronic devices and innovative electronic technologies fabrication processes. Finally, he has authored some popular scientific articles appeared on widely diffused Italian magazines and journals. Mr. Di Francia is a member of the European Technology Photovoltaic Platform (https://www.eupvplatform.org/). During his professional activity, he has coordinated several, national and European, research projects both scientifically and administratively. He is a Referee of many international journals and a member of the International Advisory Boards of several international conferences on materials science and sensors. He is actually the Head of the ENEA Photovoltaic and Sensor Applications Laboratory and a member of the Steering Committee of the Italian Association Sensors and Microsystems (AISEM).

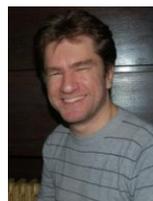

**Miloš D. Davidović** was born in Belgrade, Serbia, in 1984. He received his Dipl.Ing. (B.Sc.) and Ph.D. degrees in electrical engineering from the University of Belgrade, Serbia, in 2008 and 2015, respectively.

Currently, he is an Assistant Research Professor with the Department of Radiation and Environmental Protection at the Vinča Institute of Nuclear Sciences, University of Belgrade, Serbia. His primary research interests include air pollution modeling and monitoring, wireless sensor networks, computational physics, and signal processing. He is an author or co-author of over 50 conference presentations, papers in peer-reviewed journals, and technical reports.

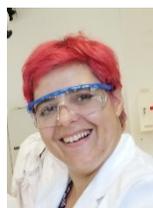

**Dr Duška Kleut** has been working at the Vinča Institute of Nuclear Sciences since 2008, one year as a holder of a PhD scholarship, and then a full time employee since 2009. She defended her PhD in electrical engineering in 2016.

Her interests are focused around two fields: Nanomaterials, Biomaterials, Materials Science, Spectroscopy, and Microscopy; and also Air quality measurements, Low-cost sensor calibration. She is currently a participant of three EU Horizon projects, VIDIS, WeBaSOOP and GrInShield.

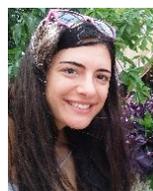

**Danka Stojanović** received her PhD from School of electrical engineering, Belgrade, Serbia. Her thesis is based on investigations and numerical simulations of electromagnetic wave propagation through chiral metamaterials. She is research associate at Center for Light-Based Research and Technologies COHERENCE, which is a part of Laboratory of atomic physics in Vinča Institute of Nuclear Sciences. Her current research interests are in development of portable biomedical devices as well as low-cost sensor technologies for air quality monitoring.

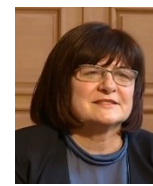

**Milena Jovašević-Stojanović** (M.S. '85, University of Belgrade, in Chemical Engineering at Faculty of Technology and Metallurgy). She received Ph.D. degree '97 in biotechnical science at University of Belgrade. From 1982 to 1998. she was researcher in respiratory protective devices (RPD) testing and development at Military Technical Institute, Belgrade investigating protective characteristic of RPD. In '98, she moved to Vinča Institute of Nuclear Sciences, Belgrade. She participated in international research projects funded by Research Council of Norway, FP6, FP7. She has been coordinator of H2020 Vidis project and Horizon Europe WeBaSOOP project. She has co-authored more than 200 scientific contributions in journals and conferences. Keywords of her research are : Air Quality, Atmospheric Aerosol Science, Human Exposure, Personal Protection. She is currently conducting research in the interdisciplinary field of air quality and its impact on human health and the environment, with a specific focus on the science of airborne particulate matter including application of PM low cost sensors and IOT wireless sensor networks. Dr. Jovašević-Stojanović has been co-chair of biannual event held in Serbia since 2007 „The International Workshop and Conference, Particulate Matter: Research and Management".

# Supplemental materials

**Supplemental Materials Table 1: Device colocation schema across the 2 seasons and 6 deployment periods.**

| # | \multicolumn | | | | | |
|---|---|---|---|---|---|---|
| | Period 1 | Period 2 | Period 3 | Period 1 | Period 2 | Period 3 |
| | Jan, 13th 15:00 -> Feb, 5 12:00; | Feb, 5th 12:00 -> Mar 2nd, 10:00; | Mar, 2nd 14:00 -> Mar 24th 10:00 | Jul, 4th 00:00 -> Jul 19th 23:59; | Aug, 24th 11:00 -> Sep, 14th 8:40; | Sep 14th 10:15 -> Oct, 4th 9:20 |
| 1 | 337 | 325 | 324 | 332 | 327 | 324 |
| 2 | 339 | 326 | 330 | 340 | 330 | 325 |
| 3 | 344 | 327 | 334 | 349 | 331 | 326 |
| 4 | 345 | 329 | 335 | 350 | 333 | 329 |
| 5 | 349 | 331 | 343 | 353 | 334 | 343 |
| 6 | 353 | 332 | 350 | 356 | 335 | _ |
| 7 | 355 | 333 | 351 | 360 | 337 | _ |
| 8 | 356 | 340 | 352 | 362 | 339 | _ |
| 9 | 360 | 341 | 362 | _ | 341 | _ |
| 10 | 361 | 364 | 363 | _ | 344 | _ |
| 11 | | | | | 345 | _ |
| 12 | _ | _ | _ | _ | 351 | _ |
| 13 | | | | | 355 | _ |
| 14 | | | | | 363 | _ |

**Supplemental Materials Table 2:** *Results of statistical significance test on the observed differences in average performance indexes (PM2.5) mean point estimation between Global calibrations (Median) models and Ad-Hoc calibrations models. Except for a few conditions for $R^2$ comparison, the test allowed to reject the null hypothesis of equal mean (although with different variance) of the indicators population and the null hypothesis of zero median in the population of observed differences pointing to a slight advantage of global calibration.*

| No. Devices | $R^2$ 2 Tails Paired t-Test | $R^2$ 2 Tails Wilcoxon Signed Rank Test | MAE 2 Tails Paired t-Test | MAE 2 Tails Wilcoxon Signed Rank Test |
|---|---|---|---|---|
| 1 | p>0.05 | p>0.05 | p<0.05 | P<0.05 |
| 2 | p<0.05 | p<0.05 | p<0.05 | p<0.05 |
| 3 | p<0.05 | p<0.05 | p<0.05 | p<0.05 |
| 4 | p<0.05 | p<0.05 | p<0.05 | p<0.05 |
| 5 | p<0.05 | p>0.05 | p<0.05 | p<0.05 |
| 6 | p<0.05 | p>0.05 | p<0.05 | p<0.05 |
| 7 | p<0.05 | p>0.05 | p<0.05 | p<0.05 |
| 8 | p<0.05 | p>0.05 | p<0.05 | p<0.05 |
| 9 | p<0.05 | p>0.05 | p<0.05 | p<0.05 |
| 10 | p<0.05 | p>0.05 | p<0.05 | p<0.05 |

**Supplemental Materials Table 3:** *Results of statistical significance test on the observed differences in average performance indexes (PM2.5) mean point estimation between Global calibrations (Aggregated) models and Ad-Hoc calibrations models. Except for a few conditions for $R^2$ comparison the test allowed to reject the null hypothesis of equal mean (respectively zero median for Wilcoxon Signed Rank Test on differences) at p=0.05 level. Note that paired t-Test were conducted under Behrens-Fisher condition (different population variance).*

| No. Devices | R^2 2 Tails Paired t-Test | R^2 2 Tails Wilcoxon Signed Rank Test | MAE 2 Tails Paired t-Test | MAE 2 Tails Wilcoxon Signed Rank Test |
|---|---|---|---|---|
| 1 | $p > 0.05$ | $p > 0.05$ | $p < 0.05$ | $P < 0.05$ |
| 2 | $p < 0.05$ | $p < 0.05$ | $p < 0.05$ | $p < 0.05$ |
| 3 | $p < 0.05$ | $p < 0.05$ | $p < 0.05$ | $p < 0.05$ |
| 4 | $p < 0.05$ | $p < 0.05$ | $p < 0.05$ | $p < 0.05$ |
| 5 | $p < 0.05$ | $p < 0.05$ | $p < 0.05$ | $p < 0.05$ |
| 6 | $p < 0.05$ | $p < 0.05$ | $p < 0.05$ | $p < 0.05$ |
| 7 | $p < 0.05$ | $p < 0.05$ | $p < 0.05$ | $p < 0.05$ |
| 8 | $p < 0.05$ | $p < 0.05$ | $p < 0.05$ | $p < 0.05$ |
| 9 | $p < 0.05$ | $p > 0.05$ | $p < 0.05$ | $p < 0.05$ |
| 10 | $p < 0.05$ | $p > 0.05$ | $p < 0.05$ | $p < 0.05$ |

**Supplemental Materials Table 4:** *Results of statistical significance tests on the observed differences in average performance indexes (PM2.5) mean point estimation between the two Global calibration methodologies. In all conditions the test failed to reject the null hypothesis of equal mean at p=0.05 level, while rejecting zero population median in the observed differences for Wilcoxon Signed Rank Test when number of involved multisensors is higher than 3. This signal the possibility for statistical significant differences between the two methods with a slight advantage of the aggregated method.*

| No. Devices | R^2 2 Tails Paired t-Test (same variance) | R^2 2 Tails Wilcoxon Signed Rank Test | MAE 2 Tails Paired t-Test (same variance) | MAE 2 Tails Wilcoxon Signed Rank Test |
|---|---|---|---|---|
| 1 | $p > 0.05$ | $p > 0.05$ | $p > 0.05$ | $p > 0.05$ |
| 2 | $p > 0.05$ | $p > 0.05$ | $p > 0.05$ | $p > 0.05$ |
| 3 | $p > 0.05$ | $p > 0.05$ | $p > 0.05$ | $p > 0.05$ |
| 4 | $p > 0.05$ | $p < 0.05$ | $p > 0.05$ | $p < 0.05$ |
| 5 | $p > 0.05$ | $p < 0.05$ | $p > 0.05$ | $p < 0.05$ |
| 6 | $p > 0.05$ | $p < 0.05$ | $p > 0.05$ | $p < 0.05$ |
| 7 | $p > 0.05$ | $p < 0.05$ | $p > 0.05$ | $p < 0.05$ |
| 8 | $p > 0.05$ | $p < 0.05$ | $p > 0.05$ | $p < 0.05$ |
| 9 | $p > 0.05$ | $p < 0.05$ | $p > 0.05$ | $p < 0.05$ |
| 10 | $p > 0.05$ | $p < 0.05$ | $p > 0.05$ | $p < 0.05$ |

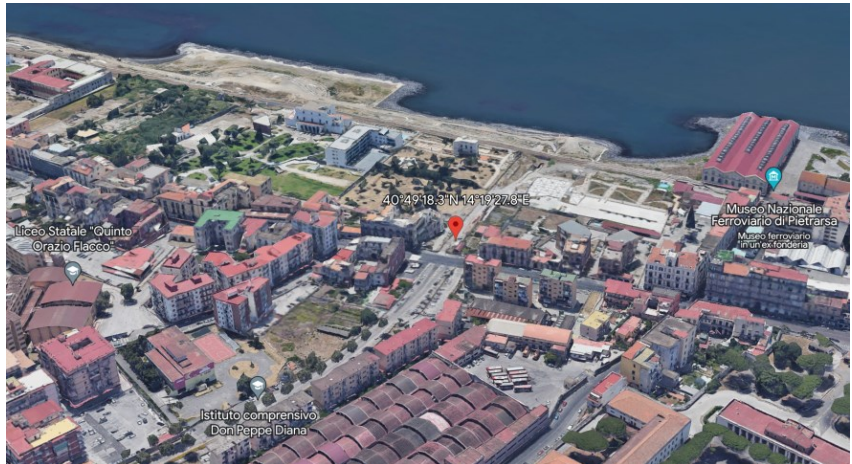
**Supplemental Materials Figure 1: Deployment Location**

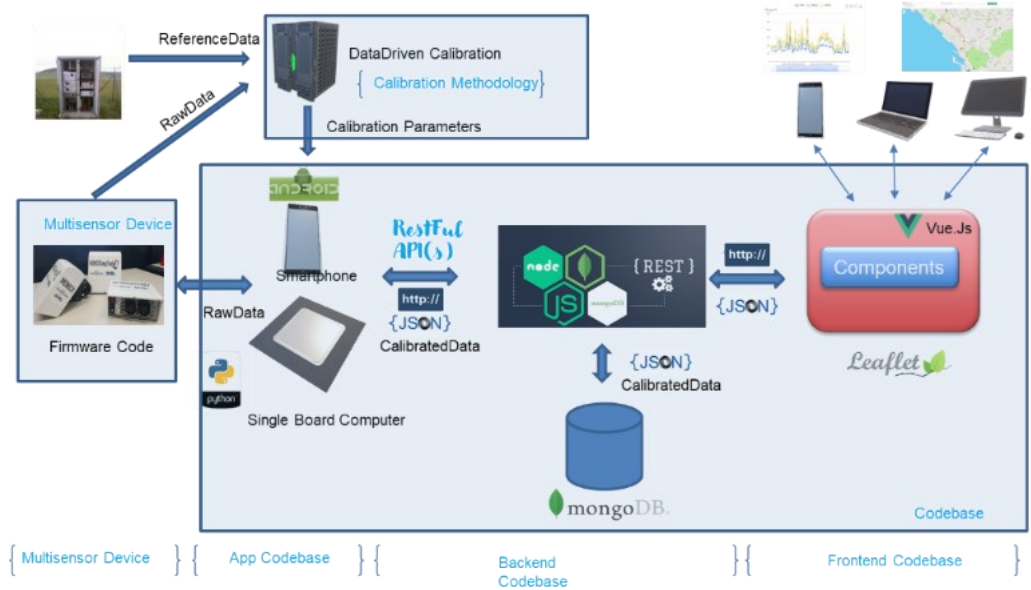
**Supplemental Materials Figure 2:**
**IoT Infrastructure implementing the generalized architecture for AirHeritage project with respect to Ingestion, Calibration, Storage and Visualization functions.**

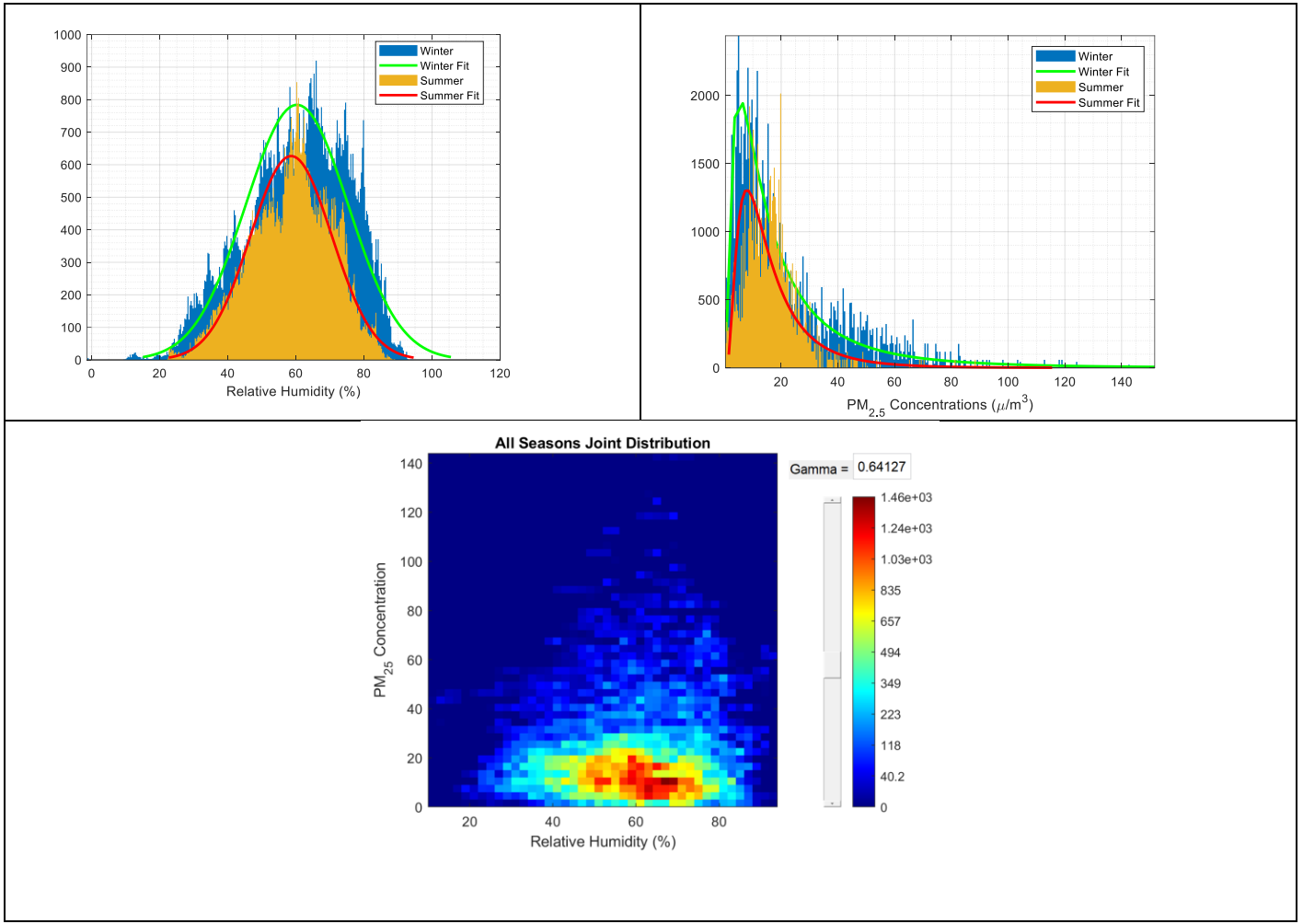

**Supplemental Materials Figure 2:** *Single (a,b) histogram visualization of recorded concentration during winter and summer colocations as fitted under the lognorm and normal distribution for, respectively, PM$_{2.5}$ pollutant concentrations and relative humidity. Joint histogram visualization (c) of both pollutant and relative humidity during the entire colocation time (Winter+Summer).*

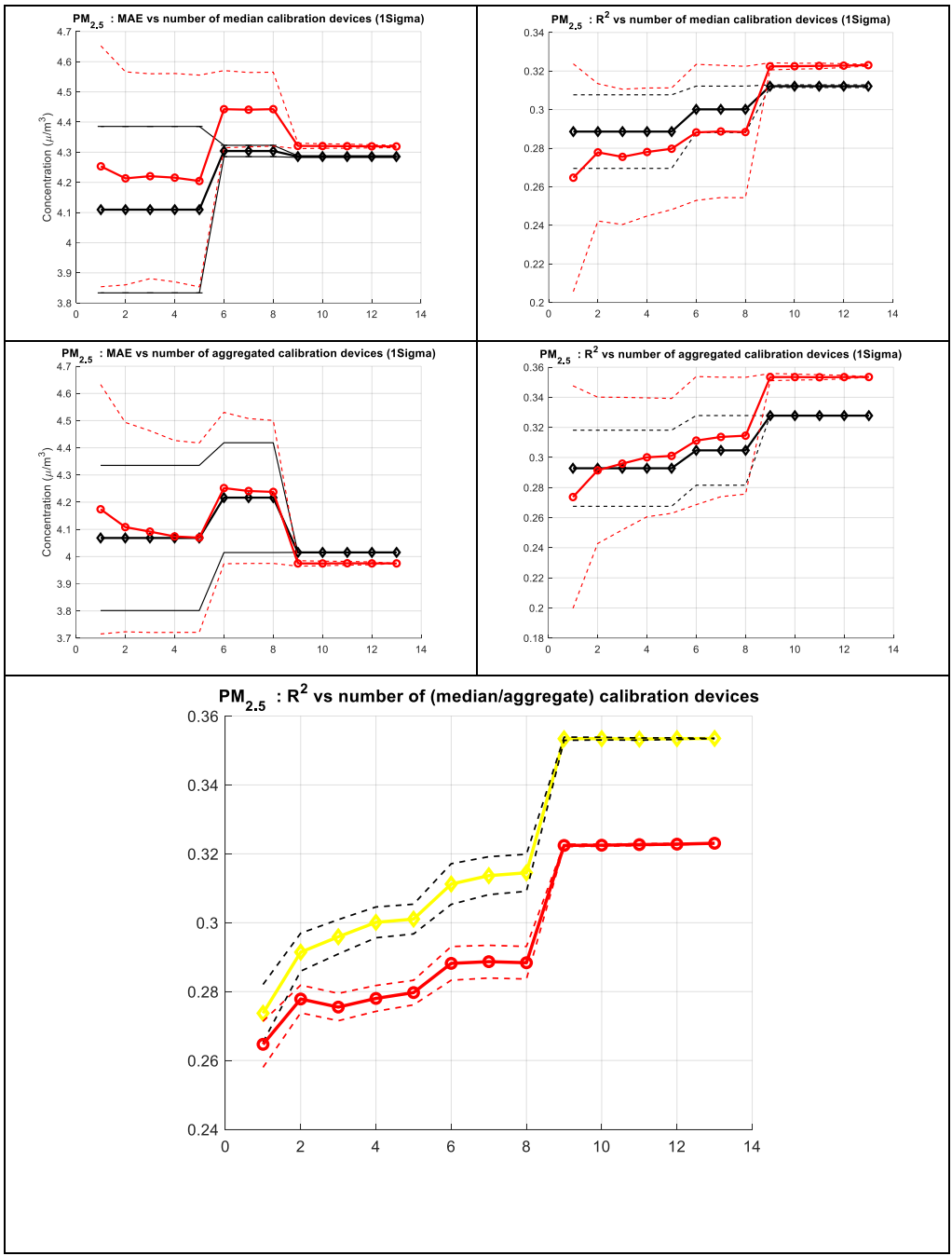

**Supplemental Materials Figure 3**: *Results obtained during summer time with short-term performance evaluation procedure (Median Approach: First row; Aggregated Approach: Second Row, Ad-Hoc performance in black in both rows). Comparison (MAE, $R^2$) between performance obtained by median (red) and aggregated (yellow) global calibration approaches during summer short-term performance evaluation. Note that, due to the peculiar deployment routine during the summer ecolocation exercise, starting from the involvement of 5 devices, the number of effective training periods is reduced to 2 (form 6 to 8 global calibration set devices) and 1 (from 9 to 13 global calibration set devices).*

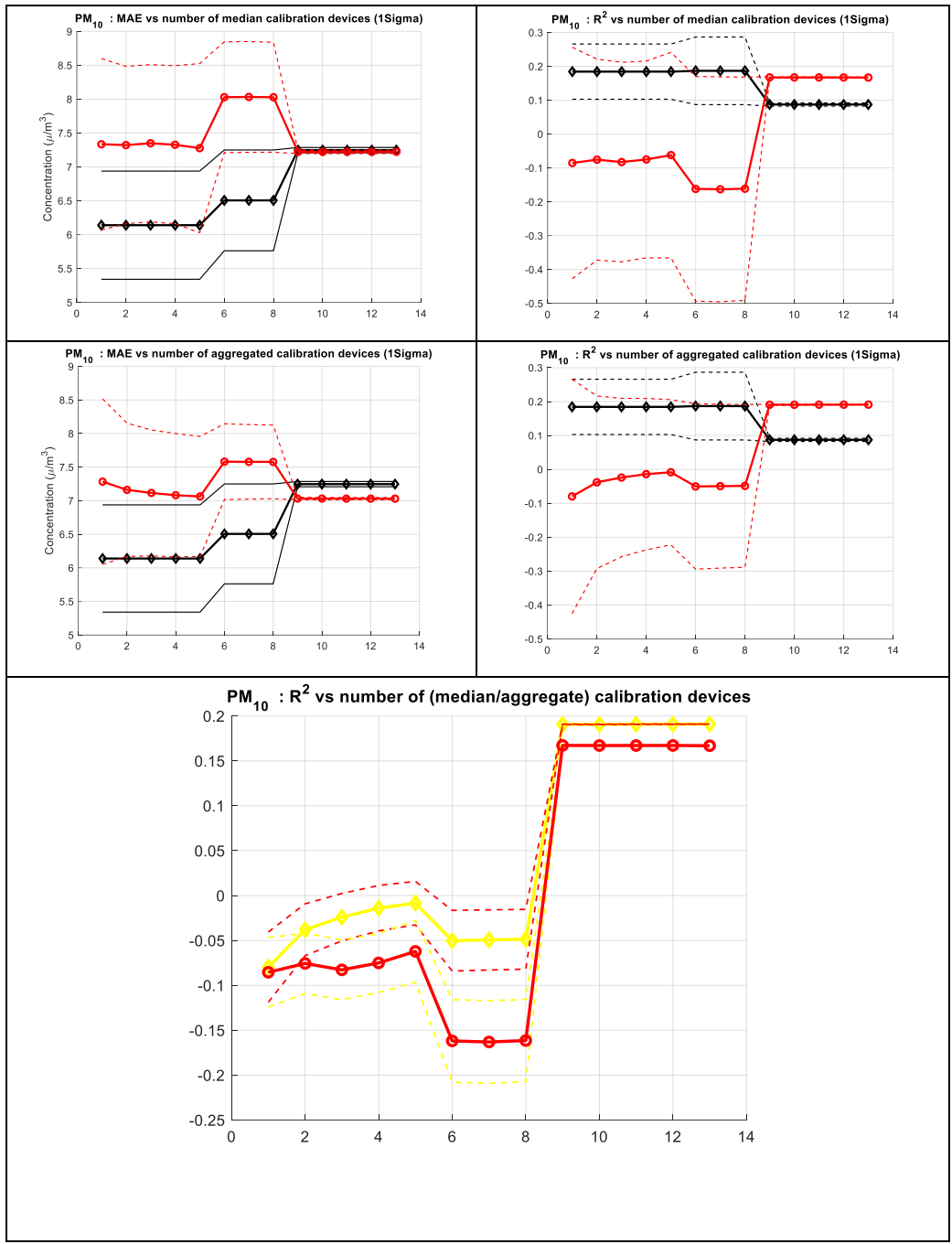

**Supplemental Materials Figure 4:** *Results obtained during summer time with short-term performance evaluation procedure (Median Approach: First row; Aggregated Approach: Second Row, Ad-Hoc performance in black in both rows). Comparison (MAE, $R^2$) between performance obtained by median (red) and aggregated (yellow) global calibration approaches during summer short-term performance evaluation. Note that, due to the peculiar deployment routine during the summer ecolocation exercise, starting from the involvement of 5 devices, the number of effective training periods is reduced to 2 (form 6 to 8 global calibration set devices) and 1 (from 9 to 13 global calibration set devices).*

```
for I = 1 to 100   // for 100 different shuffling of each period's device lists
 for period p = 1 to 3
    nlist = period p devices list
   shuffle(nlist)
   for n = 1 to  size(nlist)
      for k = 1 to n
        nodes_datasets(k,:) = extract 2 weeks of time indexed data from the nlist(k) device dataset;
      end;
      if mode = 'aggregate'
        global_training_set=nodes_dataset; // just aggregate the selected data
      else
        for each hour t in time_range(nodes_dataset)
        global_training_set(t) = median(nodes_daset(:,t));
       end;
      fit a MLR model GM over global_training_set
      for each node M in the remaining 2 periods
         divide the node M data nodeM_week_dataset in 3 weeks
         fit a MLR model AH1 over nodeM_week_dataset(1:2);
         ADP1= evaluate AH1 on nodeM_week_dataset(3);
         GM1 =evaluate GM on nodeM_week_dataset(3);
         fit a MLR model AH2 over over nodeM_week_dataset(2:3);
         ADP2 = evaluate  AH2 on nodeM_week_dataset(1);
         GMP2 = evaluate GM on nodeM_week_dataset(1);
         add avg(GMP1,GMP2) in GMPListₙ
         add avg(ADP1,ADP2) in ADPListₙ
      end;
      AHPerfList(count,n) = avg (ADPListₙ);
      GMPerfList(count,n) = avg (GMPListₙ);
   end;
    count++;
 end;
end;
```

*Supplemental Materials Inset 1: Short-term test performance computation algorithm in pseudocode. GMPerfList(i,j) is a tensor whose elements contains the vector of performance indices computed at the i-th iteration for j devices being used in the global calibration computation. Note the quantities in red could be actually separately computed, given they are independent from count and n, they are listed here for clarity.*

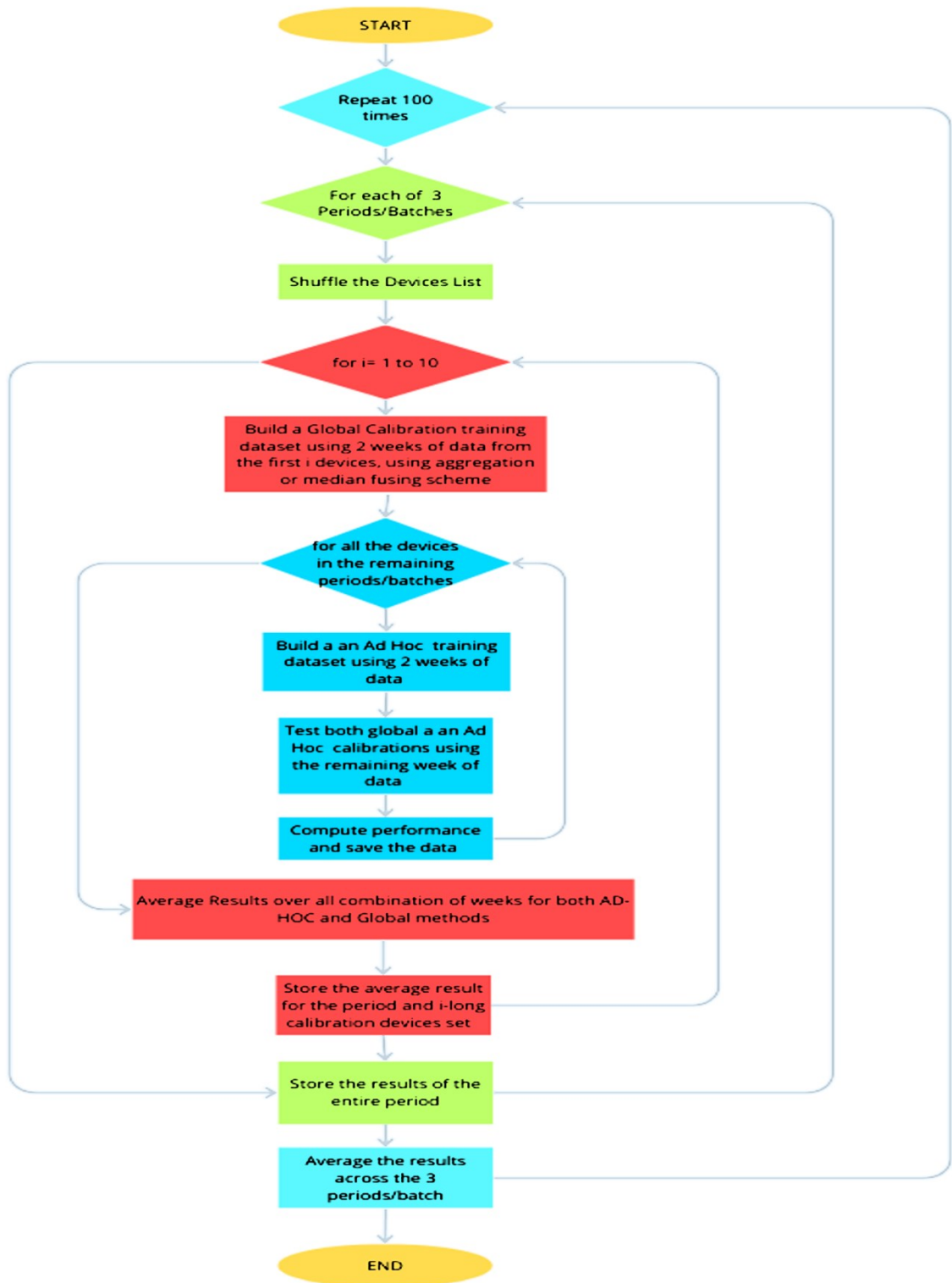

*Supplemental Materials Flowchart 1. Flowchart of short-term methodologies test performance computation algorithm*